\documentclass[runningheads]{llncs}

\usepackage{eccv}

\usepackage{eccvabbrv}

\usepackage{graphicx}
\usepackage{booktabs}
\usepackage{multirow}
\usepackage{hhline}
\usepackage{bm}
\usepackage[accsupp]{axessibility}  

\usepackage[pagebackref,breaklinks,colorlinks]{hyperref}

\usepackage{orcidlink}

\begin{document}
\begin{sloppypar}
\title{Arbitrary-Scale Video Super-Resolution with Structural and Textural Priors} 

\titlerunning{ST-AVSR}

\author{Wei Shang\inst{1,2} \and
Dongwei Ren\inst{1}\thanks{Corresponding author.} \and 
Wanying Zhang\inst{1} \and \\
Yuming Fang\inst{3} \and
Wangmeng Zuo\inst{1} \and
Kede Ma\inst{2}}

\authorrunning{W. Shang~\etal}

\institute{
School of Computer Science and Technology, Harbin Institute of Technology \\ \and 
Department of Computer Science, City University of Hong Kong  \and
Jiangxi University of Finance and Economics\\
\email{\{csweishang,rendongweihit,swzwanying\}@gmail.com \\
leo.fangyuming@foxmail.com, wmzuo@hit.edu.cn, kede.ma@cityu.edu.hk}}

\maketitle

\begin{abstract} 

Arbitrary-scale video super-resolution (AVSR) aims to enhance the resolution of video frames, potentially at various scaling factors, which presents several challenges regarding spatial detail reproduction, temporal consistency, and computational complexity. In this paper, we first describe a strong baseline for AVSR by putting together three variants of elementary building blocks: 1) a flow-guided recurrent unit that aggregates spatiotemporal information from previous frames, 2) a flow-refined cross-attention unit that selects spatiotemporal information from future frames, and 3) a hyper-upsampling unit that generates scale-aware and content-independent upsampling kernels. We then introduce ST-AVSR by equipping our baseline with a multi-scale structural and textural prior computed from the pre-trained VGG network. This prior has proven effective in discriminating structure and texture across different locations and scales, which is beneficial for AVSR. Comprehensive experiments show that ST-AVSR significantly improves super-resolution quality, generalization ability, and inference speed over the state-of-the-art. The code is available at {\url{https://github.com/shangwei5/ST-AVSR}}.

  \keywords{Arbitrary-scale video super-resolution \and Structural and textural priors}
\end{abstract}

\section{Introduction}
\label{sec:intro}
The evolutionary and developmental processes of our visual systems have presumably been shaped by continuous visual data~\cite{wandell1995foundations}. Yet, how to acquire and represent a natural scene as a continuous signal remains wide open. This difficulty stems from two main factors. The first is the physical limitations of digital imaging devices, including sensor size and density, optical diffraction, lens quality, electrical noise, and processing power. The second is the inherent complexities of natural scenes, characterized by their wide and deep frequencies, which pose significant challenges for applying the Nyquist–Shannon sampling~\cite{oppenheim1997signals} and compressed sensing~\cite{donoho2006compressed} theories to accurately reconstruct continuous signals from discrete samples. Consequently, natural scenes are predominantly represented as discrete pixel arrays, often with limited resolution.

Super-resolution (SR) provides an effective means of enhancing the resolution of low-resolution (LR) images and videos~\cite{irani1991improving,shechtman2005space}. Early deep learning-based SR methods~\cite{dong2014learning,shi2016real,lim2017enhanced,zhang2018residual} focus on fixed integer scaling factors (\eg, $\times 4$ and $\times 8$), each corresponding to an independent convolutional neural network (CNN). This limits their applicability in real-world scenarios, where varying scaling requirements are common. From the human vision perspective, users may want to continuously zoom in on images and videos to arbitrary scales using the two-finger pinch-zoom feature on mobile devices as a natural form of human-computer interaction.  From the machine vision perspective, different applications (such as computer-aided diagnosis, remote sensing, and video surveillance) may require different scaling factors to zoom in on different levels of detail for optimal analysis and decision-making.

Recently, arbitrary-scale image SR (AISR)~\cite{hu2019meta,wang2021learning,lee2022local,wang2023deep,chen2023cascaded,cao2023ciaosr} has gained significant attention due to its capability of upsampling LR images to arbitrary high-resolution (HR) using a single model. 
Contemporary AISR methods can be categorized into three classes based on how arbitrary-scale upsampling is performed: interpolation-based methods~\cite{kim2016accurate,behjati2021overnet}, learnable adaptive filter-based methods~\cite{hu2019meta,wang2021learning,wang2023deep}, and implicit neural representation-based methods~\cite{chen2021learning,lee2022local, chen2023cascaded}. These algorithms face several limitations, including quality degradation at high (and possibly integer) scales~\cite{hu2019meta,chen2021learning,wang2021learning}, high computational complexity~\cite{lee2022local, chen2023cascaded}, 
and difficulty in generalizing across unseen scales and degradation models~\cite{hu2019meta,chen2021learning,lee2022local, chen2023cascaded}, as well as temporal inconsistency in video SR.

Compared to AISR,  arbitrary-scale video SR (AVSR) is significantly more challenging due to the added time dimension.
Existing AVSR methods~\cite{chen2022videoinr,chen2023motif} rely primarily on conditional neural radiance fields~\cite{mildenhall2020nerf} as continuous signal representations. Due to the high computational demands during training and inference, only two adjacent frames are used for spatiotemporal modeling, which is bound to be suboptimal.

In this work, we aim for AVSR with the goal of reproducing faithful spatial detail and maintaining coherent temporal consistency at low computational complexity. We first describe a strong baseline, which we name B-AVSR, by identifying and combining three variants of elementary building blocks~\cite{chan2022basicvsr++,zhang2023lmr,vasconcelos2023cuf}: 1) a flow-guided recurrent unit, 2) a flow-refined cross-attention unit, and 3) a hyper-upsampling unit. The flow-guided recurrent unit captures long-term spatiotemporal dependencies from \textit{previous} frames. The flow-refined cross-attention unit first rectifies the flow estimation inaccuracy.  The refined features are then used to select beneficial spatiotemporal information from a local window of \textit{future} frames via cross-attention, which complements the flow-guided recurrent unit. The hyper-upsampling unit trains a hyper-network~\cite{ha2017hypernetworks} that takes scale-relevant parameters as input to generate content-independent upsampling kernels, enabling pre-computation to accelerate inference speed.

\begin{figure*}[!t]\footnotesize
	\centering
	\setlength{\abovecaptionskip}{3pt} 
	\setlength{\belowcaptionskip}{0pt}
	\begin{tabular}{cccccc}
		\includegraphics[width=\linewidth]{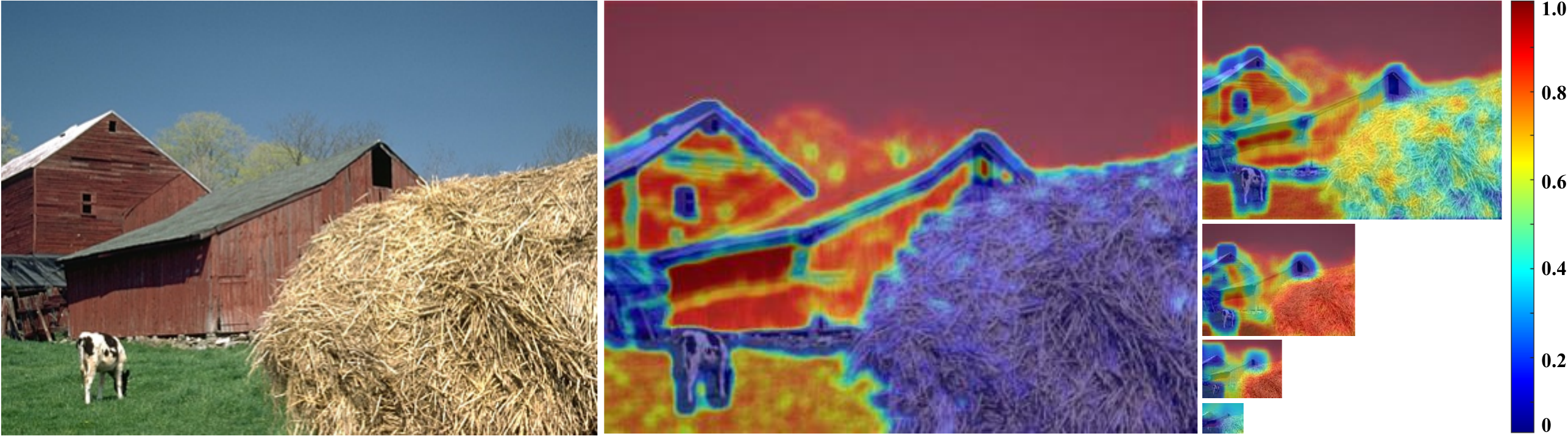}\\
	\end{tabular}
	\caption{
            Visualization of our multi-scale structural and textural prior derived from the pre-trained VGG network. A warmer color indicates a higher probability that the local patch at a given scale will be perceived as visual texture. Image borrowed from~\cite{ding2021adists} with permission.
 	}
	\label{fig:texture}
\end{figure*}
Furthermore, we introduce our complete AVSR solution, ST-AVSR, which enhances B-AVSR by incorporating a multi-scale structural and textural prior. ST-AVSR is rooted in the scale-space theory~\cite{lindeberg2013scale} in computer vision and image processing, which suggests that 
 human perception and interpretation of real-world structure and texture are scale-dependent. As shown in Fig.~\ref{fig:texture}, the hay area (located at the bottom right of the image) can be perceived alternately as structure and texture at different scales. Precisely characterizing such structure-texture transition across scales would be immensely beneficial for AVSR. Inspired by~\cite{ding2021adists}, we derive the multi-scale structural and textural prior from the multi-stage feature maps of the pre-trained VGG network~\cite{simonyan2015very}. These feature maps have proven effective in discriminating structure and texture across scales and in capturing mid-level visual concepts related to image layout~\cite{fu2023dreamsim}.

In summary, our main technical contributions include
\begin{itemize}
    \item 
    A strong baseline, B-AVSR, that is a nontrivial combination of three variants of elementary building blocks in literature~\cite{chan2022basicvsr++,zhang2023lmr,vasconcelos2023cuf},
    \item A high-performing AVSR algorithm, ST-AVSR, that leverages a powerful multi-scale structural and textural prior, and 
    \item  A comprehensive experimental demonstration, that ST-AVSR significantly surpasses competing methods in terms of SR quality on different test sets, generalization ability to unseen scales and degradation models, as well as inference speed.
\end{itemize}

\section{Related Work}

In this section, we review key components of VSR, upsampling modules for AISR and AVSR, and natural scene priors employed in SR.
\subsection{Key Components of VSR}
 Kappeler \etal~\cite{kappeler2016video} pioneered CNN-based approaches for VSR, emphasizing two key components: feature alignment and aggregation. Subsequent studies have focused on enhancing these components. EDVR~\cite{wang2019edvr} introduced pyramid deformable alignment and spatiotemporal attention for feature alignment and aggregation.
 BasicVSR~\cite{chan2021basicvsr} and BasicVSR$++$~\cite{chan2022basicvsr++}
employ an optical flow-based module to estimate motion correspondence between neighboring frames for feature alignment and a bidirectional propagation module to aggregate spatiotemporal information from previous and future frames, which set the VSR performance record at that time.
RVRT~\cite{liang2022recurrent} enhanced VSR performance by utilizing a recurrent video restoration Transformer with guided deformable attention albeit at the expense of substantially increased computational complexity.
Additionally, MoTIF~\cite{chen2023motif}  integrated VSR with video frame interpolation, which achieved limited success due to the ill-posedness of the task.
In our work, we combine a flow-guided recurrent unit and a flow-refined cross-attention unit to extract, align, and aggregate spatiotemporal features from previous and future frames, while keeping computational complexity manageable.

\subsection{Upsampling Modules for AISR and AVSR}

Compared to fixed-scale SR methods ~\cite{dong2014learning,shi2016real,lim2017enhanced,zhang2018residual,liang2021swinir}, upsampling plays a more crucial role in AISR and AVSR.
Besides direct interpolation-based upsampling~\cite{kim2016accurate,behjati2021overnet}, learnable adaptive filter-based upsampling and implicit neural representation-based upsampling are commonly used. Meta-SR~\cite{hu2019meta} was the pioneer in AISR, dynamically predicting the upsampling kernels using a single model.
ArbSR~\cite{wang2021learning} introduced a scale-aware upsampling layer compatible with fixed-scale SR methods.
EQSR~\cite{wang2023deep} proposed a bilateral encoding of both scale-aware and content-dependent features during upsampling.
Inspired by the success of implicit neural representations in computer graphics~\cite{michalkiewicz2019implicit,peng2020convolutional}, this approach has also been applied to AISR and AVSR.
For instance, LIIF~\cite{chen2021learning} predicts the RGB values of HR pixels using the coordinates of LR pixels along with their neighboring features as inputs.
LTE~\cite{lee2022local} captures more fine detail with a local texture estimator, and
CLIT~\cite{chen2023cascaded} enhances representation expressiveness 
with cross-scale attention and multi-scale reconstruction.
OPE~\cite{song2023ope} introduced orthogonal position encoding for efficient upsampling. 
CiaoSR~\cite{cao2023ciaosr} proposed an attention-based weight ensemble algorithm for feature aggregation in a large receptive field.

Existing AVSR methods~\cite{chen2022videoinr,chen2023motif} also use implicit neural representations but are constrained to modeling spatiotemporal relationships between only two adjacent frames due to the high computational costs involved. 
The proposed ST-AVSR addresses this limitation by employing a lightweight hyper-upsampling unit to predict scale-aware and content-independent upsampling kernels, allowing for pre-computation to speed up inference.

\subsection{Natural Scene Priors for SR}
The history of SR, or more generally low-level vision, is closely tied to the development of natural scene priors. Commonly used priors in SR include the smoothness prior~\cite{chambolle2004algorithm}, sparsity prior~\cite{mairal2014sparse}, self-similarity prior~\cite{glasner2009super}, edge/gradient prior~\cite{he2011single}, deep architectural prior~\cite{ulyanov2018deep}, temporal consistency prior~\cite{caballero2017real}, motion prior~\cite{tao2017detail,shang2023joint}, and perceptual prior~\cite{wang2018recovering}. In the subfield of AISR and AVSR, the scaling factor-based priors have exclusively been leveraged~\cite{wang2021learning,fu2021residual,wang2023deep}. In this paper, we introduce a multi-scale structural and textural prior that effectively separates structure and texture at varying locations and scales,  capturing their alternating and smooth transitions. We demonstrate its effectiveness in enhancing AVSR.

\begin{figure*}[!t]\footnotesize
	\centering
	\setlength{\abovecaptionskip}{3pt} 
	\setlength{\belowcaptionskip}{0pt}
	\begin{tabular}{cccccc}
		\includegraphics[width=\linewidth]{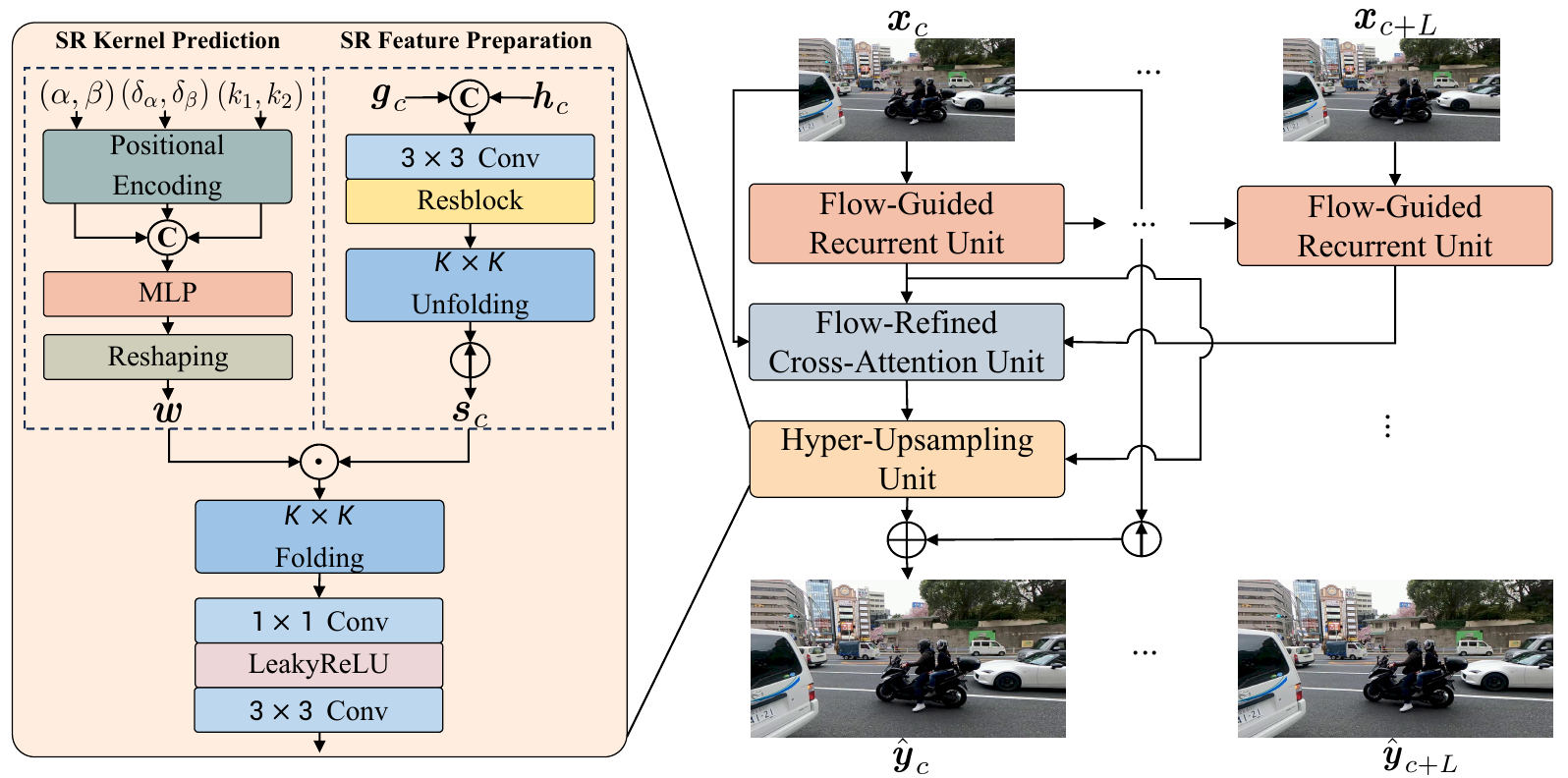}\\
	\end{tabular}
	\caption{System diagram of B-AVSR, which reconstructs an arbitrary-scale HR video $\hat{\bm y}$ from an LR video input $\bm x$. B-AVSR is composed of three variants of elementary building blocks: 1) a flow-guided recurrent unit to aggregate features from previous frames, 2) a flow-refined cross-attention unit to select features from future frames (see also Fig.~\ref{fig:local}), and 3) a hyper-upsampling unit to prepare SR features and predict SR kernels for HR frame reconstruction. ST-AVSR is built on top of B-AVSR by replacing all instances of $\bm x$ with the multi-scale structural and textural prior $\bm p$ (see the detailed text description in Sec.~\ref{subsec:st}).
	}
	\label{fig:framework}
\end{figure*}

\section{Proposed Method: ST-AVSR}
Given an LR video sequence $\bm{x} = \{\bm{x}_{i}\}_{i=0}^T$, where $\bm x_{i}\in\mathbb{R}^{H\times W}$ is the $i$-th frame, and $H$ and $W$ are the frame height and width, respectively, the goal of the proposed B-AVSR and ST-AVSR is to reconstruct an HR video sequence $\hat{\bm{y}} = \{\hat{\bm{y}}_{i}\}_{i=0}^T$ with $\hat{\bm y}_{i} \in \mathbb{R}^{(\alpha H)\times (\beta W)}$, where $\alpha,\beta\ge 1$ are two user-specified scaling factors. Our baseline B-AVSR consists of three variants of basic building blocks: 1) a flow-guided recurrent unit,
2) a flow-refined cross-attention unit, and 3) a hyper-upsampling unit. ST-AVSR enhances B-AVSR by incorporating a multi-scale structural and textural prior. The system diagram is shown in Fig.~\ref{fig:framework}.

\subsection{Flow-Guided Recurrent Unit}
Given $\bm x = \{\bm x_{i}\}_{i=0}^T$, the flow-guided recurrent unit computes a sequence of hidden states $\{\bm h_{i}\}_{i=1}^{T}$ to capture long-term spatiotemporal dependencies of previous frames. Initially, we estimate the optical flow between the current and previous frames~\cite{chan2021basicvsr}:
\begin{align}
    \bm{f}_{i\rightarrow{i-1}} = \mathtt{flow}\left(\bm{x}_{i}, \bm{x}_{i-1}\right), \quad i \in \{1,2,\ldots, T\},
\end{align}
where $\mathtt{flow}(\cdot)$ denotes a state-of-the-art optical flow estimator~\cite{sun2018pwc}. 
$\bm{f}_{i\rightarrow{i-1}}$ is then used to align the hidden state $\bm h_{i-1}$ backward:
\begin{align}
    \bm h_{i-1\rightarrow i} = \mathtt{warp}(\bm h_{i-1}, \bm f_{i \rightarrow i-1}), \quad i \in \{1,2,\ldots, T\},
\end{align}
where $\mathtt{warp}(\cdot)$ denotes the standard image/feature warping operation using the bilinear kernel, and $\bm h_{0} = \bm 0$. Subsequently, the aligned previous hidden state $\bm h_{i-1\rightarrow i}$ and the current frame $\bm x_{i}$ are concatenated along the channel dimension and processed through a ResNet with $N_1$ residual blocks to compute $\bm h_i$.
The flow-guided recurrent unit allows the proposed B-AVSR to incorporate long-term historical context while being flow-aware.
\begin{figure*}[!t]\footnotesize
	\centering
	\setlength{\abovecaptionskip}{3pt} 
	\setlength{\belowcaptionskip}{0pt}
	\begin{tabular}{cccccc}
		\includegraphics[width=0.9\linewidth]{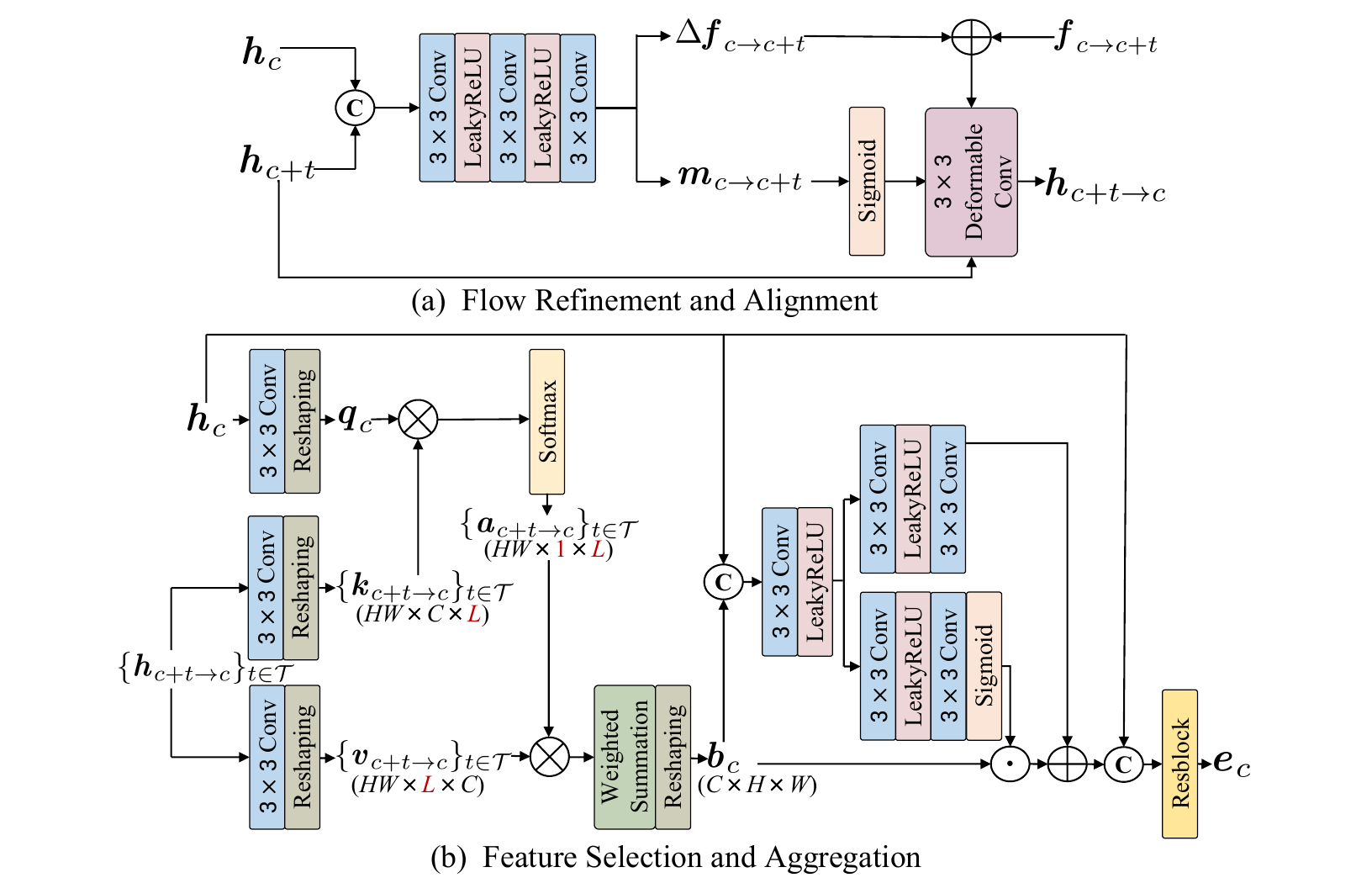}\\
	\end{tabular}
	\caption{Computational structure of the flow-refined cross-attention unit.
	}
	\label{fig:local}
\end{figure*}

\subsection{Flow-Refined Cross-Attention Unit} 
The computation of $\bm h_i$ in the flow-guided recurrent unit depends entirely on features extracted from previous frames. To benefit from future frames, similar to bidirectional recurrent networks, but without the need for computing and storing backward hidden states, we use a sliding window approach to selectively aggregate spatiotemporal information from $L$ future frames. Specifically,
given a local future window of frames $\{\bm x_{i}\}_{i=c}^{c+L}$, 
we first compensate for the inaccuracy in optical flow estimation between $\bm x_{c}$ and $\bm x_{c+t}$ due to their potentially large temporal interval~\cite{chan2022basicvsr++}. As shown in Fig.~\ref{fig:local} (a), we adopt a lightweight CNN with three convolution layers and LeakyReLU activations in between to predict the flow offsets $\Delta \bm f_{c\rightarrow c+t}$ and modulation scalars $\bm m_{c\rightarrow c+t}$:
\begin{align}\label{eq:fr1}
     \Delta\bm{f}_{c\rightarrow{c+t}},\bm{m}_{c\rightarrow{c+t}} &= \mathtt{SConv}(\bm{h}_{c}, \bm{h}_{c+t}), \quad t\in \mathcal{T} =\{1,\ldots, L\},
\end{align}
where $\mathtt{SConv}(\cdot)$ denotes a generic CNN with standard convolutions. We then rectify the flow estimation as $\bm{f}_{c\rightarrow{c+t}} +\Delta\bm{f}_{c\rightarrow{c+t}} $, where $\bm{f}_{c\rightarrow{c+t}} = \mathtt{flow}(\bm x_{c},\bm x_{c+t})$ and use it together with $\bm{m}_{c\rightarrow{c+t}}$ to align $\bm h_{c+t}$:
\begin{align}\label{eq:fr2}
    \bm{h}_{{c+t}\rightarrow c} = \mathtt{DConv}(\bm{h}_{c+t}, \bm{f}_{c\rightarrow{c+t}} +\Delta\bm{f}_{c\rightarrow{c+t}} , \mathtt{sigmoid}(\bm{m}_{c\rightarrow{c+t}})),
\end{align}
where we normalize the modulation scalars as $\mathtt{sigmoid}(\bm{m}_{c\rightarrow{c+t}})$. $\mathtt{DConv}(\cdot)$ denotes a generic CNN with modulated deformable convolutions~\cite{zhu2019deformable}. Here $\mathtt{DConv}(\cdot)$  is implemented by a single deformable convolutional layer. 

We selectively aggregate useful future information via \textit{local} cross-attention. As shown in Fig.~\ref{fig:local} (b), the query $\bm{q}_{c}$ is derived from the current hidden state $\bm{h}_{c}$. The key $\bm k_{c+t\rightarrow 
 c}$ and the value $\bm v_{c+t\rightarrow 
 c}$ are generated from the $t$-th aligned hidden state $\bm{h}_{c+t\rightarrow c}$, where $t\in\mathcal{T}=\{1,\ldots, L\}$:
\begin{align}\label{eq:fr3}
     \bm{q}_{c} = \mathtt{SConv}(\bm{h}_{c}), \bm{k}_{c+t\rightarrow 
 c} = \mathtt{SConv}(\bm{h}_{c+t\rightarrow 
 c}), \, \mathrm{and} \, \bm{v}_{c+t\rightarrow 
 c} = \mathtt{SConv}(\bm{h}_{c+t\rightarrow c}).
\end{align}
We measure the feature similarity between the query $\bm{q}_{c}(z)$ and the keys $\{\bm k_{c+t\rightarrow 
 c}(z)\}_{t\in\mathcal{T}}$ at spatial position $z$ using inner product $\left\langle\cdot,\cdot\right\rangle$:
\begin{align}
\bm{a}_{c+t\rightarrow 
 c}(z) = \frac{\exp \left(\left\langle \bm{q}_{c}(z), \bm{k}_{c+t\rightarrow 
 c}(z)\right\rangle\right)}{\sum_{t'\in \mathcal{T}} \exp \left(\left\langle \bm{q}_{c}(z), \bm{k}_{c+t'\rightarrow 
 c}(z)\right\rangle\right)},
\end{align}
where $\bm a_{c+t\rightarrow 
 c}$ is the $t$-th attention map. The aggregated features from the $L$ future frames can be computed by  a weighted summation:
\begin{equation}\label{eq:fr4}
\bm{b}_c(z) = \sum_{t\in\mathcal{T}} \bm{a}_{c+t\rightarrow 
 c}(z)\cdot \bm{v}_{c+t\rightarrow 
 c}(z).
\end{equation}
To further enhance feature selection and aggregation, we implement a variant of the squeeze and excitation mechanism~\cite{hu2018squeeze} as a form of \textit{global} self-attention. It computes the enhanced features $\bm d_c$ from $\bm b_c$ with reference to $\bm h_c$: 
\begin{align}\label{eq:fr5}
    \bm d_c = \mathtt{SConv}(\bm h_{c}, \bm b_c) + \bm b_c\odot\mathtt{sigmoid}\left(\mathtt{SConv}(\bm h_{c}, \bm b_c)\right).
\end{align}
Next, we merge the current hidden state $\bm h_{c}$ with
 $\bm d_c$:
\begin{align}\label{eq:fr6}
\bm e_c = \mathtt{SConv}(\bm h_{c}, \bm d_c),
\end{align}
and concatenate it with the current frame $\bm x_{c}$ along the channel dimension to compute the final output features $\bm{g}_c$ through a ResNet with $N_2$ residual blocks.

\subsection{Hyper-Upsampling Unit}
Inspired by the neural kriging upsampler~\cite{wang2023deep}, our hyper-upsampling unit consists of two branches: SR feature preparation and SR kernel prediction, as shown in Fig.~\ref{fig:framework}. For SR feature preparation, we concatenate the output features $\bm{g}_c$ from the flow-refined cross-attention unit with the current hidden state $\bm{h}_{c}$, and pass them through a residual block to compute SR features.
Next, we unfold a $K\times K$ spatial neighborhood of $C$-dimensional SR feature representations into $C\times K^2$ channels (\ie, the tensor generalization of $\mathtt{img2col}(\cdot)$ in image processing). Finally, we upsample the unfolded features to the target resolution using bilinear interpolation, resulting in $\bm{s}_{c}$.

For SR kernel generation, we train a hyper-network, \ie, a
 multi-layer perceptron (MLP) with periodic activation functions~\cite{chen2023cascaded}, to predict the upsampling kernels $\bm w$.  
 Periodic activations have been shown to effectively address the spectral bias of MLPs, outperforming ReLU non-linearity~\cite{sitzmann2020implicit}. The inputs to the MLP are carefully selected to be scale-aware and content-independent. These include
 1) the scaling factors $(\alpha, \beta)$, 2) the relative coordinates between the LR and HR frames $(\delta_\alpha,\delta_\beta)$, and 3) the spatial indices $(k_1, k_2)$ of $\bm w$. The first two inputs have been used in other continuous representation methods~\cite{chen2021learning,lee2022local}.
To enhance the discriminability of scale-relevant inputs, we employ sinusoidal positional encoding as a pre-processing step.
It is noteworthy that our upsampling kernels $\bm w$ can be pre-computed and stored for various target resolutions, which accelerates inference time. 

After obtaining $\bm w$, we perform Hadamard multiplication between $\bm w$ and $\bm{s}_{c}$, followed by a folding operation (\ie, the inverse of the unfolding operation).
Finally, we employ a $1\times 1$ convolution to blend information across the channel dimension, followed by a $3\times 3$ convolution for channel adjustment, with LeakyReLU in between. The output from the last $3\times 3$ convolution layer is then added to the upsampled LR frame to produce the final HR frame, $\hat{\bm y}_c$.

\subsection{Multi-Scale Structural and Textural Priors for AVSR}\label{subsec:st}
Accurately characterizing image structure and texture at multiple scales is crucial for the task of AVSR. Fortunately, the scale-space theory in computer vision and image processing~\cite{koenderink1984structure,lindeberg2013scale} provides an elegant theoretical framework for this purpose. The most common approach to creating a scale space is to convolve the original image with a \textit{linear} Gaussian kernel of varying widths, using the standard deviation $\sigma$ as the scale parameter~\cite{huxley1997gaussian}. Additionally, the Laplacian of Gaussian and the difference of Gaussians are also frequently employed as linear scale-space representations, such as in the development of the influential SIFT image descriptor~\cite{lowe2004distinctive}. With the rise of deep learning, \textit{non-linear} scale-space representations have become more accessible thanks to the alternating convolution and subsampling operations in CNNs. A notable example is due to Ding~\etal~\cite{ding2021adists}, who observed that the multi-stage feature maps computed from the pre-trained VGG network~\cite{simonyan2015very} effectively discriminate structure and texture at different locations and scales, as illustrated in Fig.~\ref{fig:texture}. 
Motivates by these theoretical and computational studies, we also choose to work with the multi-stage VGG feature maps, upsampling and concatenating them along the channel dimension. Next, we apply a $1\times 1$ convolution to reduce the number of channels to $C$ and concatenate them with the current frame $\bm x_c$, which serves as the multi-scale structural and textural prior, denoted by $\bm p_c$. Inserting these structural and textural priors into the baseline model B-AVSR  is straightforward: we replace all instances of $\bm x$ with $\bm p$ (except for the last residual connection which produces the HR video $\hat{\bm y}$). This completes our ultimate AVSR model, ST-AVSR.

\begin{table}[!t]
\centering
\setlength{\abovecaptionskip}{0pt} 
\setlength{\belowcaptionskip}{0pt}
\caption{Quantitative comparison with state-of-the-art methods on the REDS validation set (PSNR$\uparrow$ / SSIM$\uparrow$ / LPIPS$\downarrow$). The best results are highlighted in boldface. }
\label{table:reds}
\renewcommand{\arraystretch}{1.2}
\resizebox{\columnwidth}{!}{
\begin{tabular}{cc|c|c|c|c|c}
\multicolumn{2}{c|}{Method} & \multicolumn{5}{c}{Scale} \\  
\hline
\multirow{2}{*}{Backbone} & Upsampling  & \multirow{2}{*}{$\times 2$} & \multirow{2}{*}{$\times 3$} & \multirow{2}{*}{$\times 4$} & \multirow{2}{*}{$\times 6$} & \multirow{2}{*}{$\times 8$}  \\ 
&  Unit & & & & & \\ \hline   
\multicolumn{2}{c|}{Bicubic} &  {\fontsize{8}{10}\selectfont 31.51/0.911/0.165}   & {\fontsize{8}{10}\selectfont 26.82/0.788/0.377 }  & {\fontsize{8}{10}\selectfont  24.92/0.713/0.484}  & {\fontsize{8}{10}\selectfont  22.89/0.622/0.631}  &  {\fontsize{8}{10}\selectfont 21.69/0.574/0.699 }  \\ 
\multicolumn{2}{c|}{EDVR~\cite{wang2019edvr}} &  {\fontsize{8}{10}\selectfont 36.03/0.961/0.072}   & {\fontsize{8}{10}\selectfont 32.59/0.904/0.108 }  & {\fontsize{8}{10}\selectfont  30.24/0.853/0.202}  & {\fontsize{8}{10}\selectfont  27.02/0.733/0.349}  &  {\fontsize{8}{10}\selectfont 25.38/0.678/0.411 }  \\
\multicolumn{2}{c|}{ArbSR~\cite{wang2021learning}} & {\fontsize{8}{10}\selectfont 34.48/0.942/0.096 } & {\fontsize{8}{10}\selectfont 30.51/0.862/0.200 } &  {\fontsize{8}{10}\selectfont 28.38/0.799/0.295 } & {\fontsize{8}{10}\selectfont 26.32/0.710/0.428} & {\fontsize{8}{10}\selectfont 25.08/0.641/0.492 }  \\
\multicolumn{2}{c|}{EQSR~\cite{wang2023deep}}   & {\fontsize{8}{10}\selectfont 34.71/0.943/0.082 }  & {\fontsize{8}{10}\selectfont 30.71/0.867/0.194}    & {\fontsize{8}{10}\selectfont 28.75/0.804/0.283 }        & {\fontsize{8}{10}\selectfont 26.53/0.718/0.391}      & {\fontsize{8}{10}\selectfont 25.23/0.645/0.459}    \\ \hline 
\multirow{3}{*}{RDN~\cite{zhang2018residual}} & LTE~\cite{lee2022local} & {\fontsize{8}{10}\selectfont 34.63/0.942/0.093 }  & {\fontsize{8}{10}\selectfont 30.64/0.865/0.204}     & {\fontsize{8}{10}\selectfont 28.65/0.801/0.289}     & {\fontsize{8}{10}\selectfont 26.46/0.714/0.410}    & {\fontsize{8}{10}\selectfont 25.15/0.660/0.488 } \\ 
& CLIT~\cite{chen2023cascaded} & {\fontsize{8}{10}\selectfont 34.63/0.942/0.092 }   & {\fontsize{8}{10}\selectfont 30.63/0.865/0.204}       & {\fontsize{8}{10}\selectfont 28.63/0.801/0.290}           & {\fontsize{8}{10}\selectfont  26.43/0.714/0.400 }    & {\fontsize{8}{10}\selectfont 25.14/0.661/0.467  }\\ 
& OPE~\cite{song2023ope} & {\fontsize{8}{10}\selectfont 34.05/0.939/0.082}   & {\fontsize{8}{10}\selectfont 30.52/0.864/0.199 }    & {\fontsize{8}{10}\selectfont 28.63/0.800/0.293 }   & {\fontsize{8}{10}\selectfont 26.37/0.711/0.421 }    & {\fontsize{8}{10}\selectfont 25.04/0.655/0.504} \\ \hline 
\multirow{3}{*}{SwinIR~\cite{liang2021swinir}} & LTE~\cite{lee2022local} & {\fontsize{8}{10}\selectfont 34.73/0.943/0.091 }      & {\fontsize{8}{10}\selectfont 30.73/0.866/0.200}     & {\fontsize{8}{10}\selectfont 28.75/0.804/0.284}      & {\fontsize{8}{10}\selectfont 26.56/0.718/0.403}         & {\fontsize{8}{10}\selectfont 25.24/0.669/0.480 } \\ 
& CLIT~\cite{chen2023cascaded} & {\fontsize{8}{10}\selectfont 34.63/0.942/0.093}       & {\fontsize{8}{10}\selectfont 30.64/0.865/0.205 }      & {\fontsize{8}{10}\selectfont 28.64/0.802/0.291}     & {\fontsize{8}{10}\selectfont 26.45/0.715/0.400}     & {\fontsize{8}{10}\selectfont 25.15/0.662/0.466} \\ 
& OPE~\cite{song2023ope} & {\fontsize{8}{10}\selectfont  33.39/0.935/0.081 }     & {\fontsize{8}{10}\selectfont 29.40/0.820/0.217 }   & {\fontsize{8}{10}\selectfont 28.49/0.785/0.292 }          & {\fontsize{8}{10}\selectfont 26.30/0.698/0.398 }    & {\fontsize{8}{10}\selectfont 25.01/0.648/0.487}  \\ \hline 

\multicolumn{2}{c|}{VideoINR~\cite{chen2022videoinr}}    &  {\fontsize{8}{10}\selectfont 31.59/0.900/0.144 }  &  {\fontsize{8}{10}\selectfont 30.04/0.852/0.197 }    & {\fontsize{8}{10}\selectfont 28.13/0.791/0.263 }   & {\fontsize{8}{10}\selectfont  25.27/0.687/0.374 } & {\fontsize{8}{10}\selectfont 23.46/0.619/0.470 }  \\ 
\multicolumn{2}{c|}{MoTIF~\cite{chen2023motif}}      & {\fontsize{8}{10}\selectfont 31.03/0.898/0.100 }      & {\fontsize{8}{10}\selectfont 30.44/0.862/0.186  }  & {\fontsize{8}{10}\selectfont 28.77/0.807/0.260 }    & {\fontsize{8}{10}\selectfont 25.63/0.698/0.369 }    & {\fontsize{8}{10}\selectfont 25.12/0.664/0.467 }     \\ \hline 
\multicolumn{2}{c|}{ST-AVSR (Ours)}     & {\fontsize{7}{9}\selectfont \textbf{36.91}/\textbf{0.969}/\textbf{0.041}}    & {\fontsize{7}{9}\selectfont  \textbf{33.41}/\textbf{0.937}/\textbf{0.066}}       & {\fontsize{7}{9}\selectfont  \textbf{31.03}/\textbf{0.897}/\textbf{0.114}}      & {\fontsize{7}{9}\selectfont \textbf{27.89}/\textbf{0.812}/\textbf{0.222} }    & {\fontsize{7}{9}\selectfont \textbf{26.04}/\textbf{0.746}/\textbf{0.298}}     \\ 
\end{tabular}}
\end{table}

\begin{figure*}[!t]\footnotesize
	\centering
	\setlength{\abovecaptionskip}{3pt} 
	\setlength{\belowcaptionskip}{0pt}
	\begin{tabular}{cccccc}
		\includegraphics[width=\linewidth]{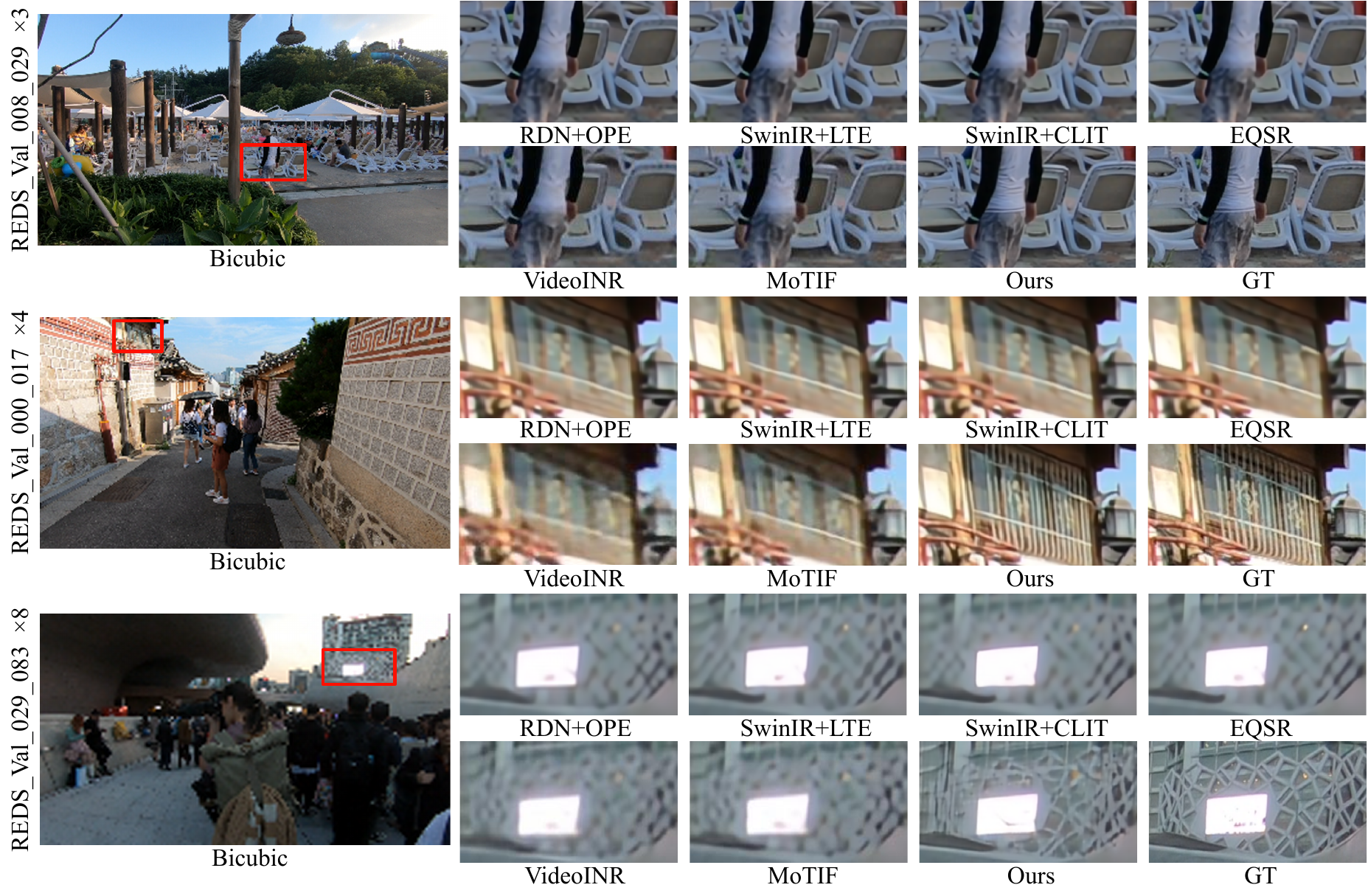}\\
	\end{tabular}
	\caption{
		Visual comparison of different AVSR methods on the REDS dataset. Zoom in for better distortion visibility.
	}
	\label{fig:reds}
\end{figure*}

\section{Experiments}
In this section, we first describe the experimental setups and then compare the proposed ST-AVSR against state-of-the-art AISR and AVSR methods, followed by a series of ablation studies to justify the key design choices of ST-AVSR, especially the incorporation of the multi-scale structural and textural prior.
\subsection{Experimental Setups}
\subsubsection{Datasets.}
ST-AVSR is trained on the REDS dataset~\cite{nah2019ntire}, which comprises $240$ videos of resolution $720\times1,280$ captured by GoPro. Each video consists of $100$ HR frames. 
Following the settings in~\cite{chen2022videoinr,chen2023cascaded,chen2023motif}, we generate LR frames using the bicubic degradation model, with randomly sampled scaling factors $(\alpha, \beta)$ from a uniform distribution $\mathcal{U}[1, 4]$. We test ST-AVSR on the validation set of REDS comprising $30$ videos, and the Vid4 dataset~\cite{liu2013bayesian} containing $4$ videos.
\subsubsection{Data Pre-processing.}
To enable mini-batch training with varying LR/HR resolutions, we adapt the pre-processing method used for AISR in EQSR~\cite{wang2023deep} to AVSR.
Specifically, from an HR video patch of size $\alpha P\times \beta P\times T$, we generate the input LR video patch by \textit{resizing} it to $P\times P\times T$. We next \textit{crop} a set of ground-truth patches of size  $P\times P\times T$  from the same HR patch. The respective relative coordinates $(\delta_\alpha, \delta_\beta)$ are recorded for use in the hyper-upsampling unit to differentiate between different ground-truth patches for the same input (see also the data pre-processing pipeline in the Supplementary). Data augmentation techniques include random rotation (by $90^\circ$, $180^\circ$, or $270^\circ$) and random horizontal and vertical flipping.

\subsubsection{Implementation Details.}
ST-AVSR is end-to-end optimized for $300$K iterations. Adam~\cite{kingma2014adam} is chosen as the optimizer, with an initial learning rate $2\times$10$^{-4}$ 
that is gradually lowered to 1$\times$10$^{-6}$ by cosine annealing~\cite{loshchilov2016sgdr}. 
We set the input patch size to $P=80$, the sequence length to $T=15$, the sliding window size to $L=2$,  the number of ResBlocks to $N_1=N_2=15$, the unfolding neighborhood to $K=3$, and the SR feature dimension to $C=64$, respectively.
The hidden dimensions of the MLP in the hyper-upsampling unit are $16$, $16$, $16$, and $64$, respectively.
The parameters of PWC-Net~\cite{sun2018pwc} as the optical flow estimator and the pre-trained VGG network to derive the multi-scale structural and textural prior are frozen during training.
We use the Charbonnier loss~\cite{lai2017deep}:
\begin{equation}
   \ell(\hat{\bm y}, \bm y) = \frac{1}{(T+1)\vert\mathcal{Z}\vert} \sum^{T}_{i=0}\sum_{z\in\mathcal{Z}} \sqrt{ (\hat{\bm y}_{i}(z) - \bm{y}_{i}(z))^2 + \epsilon},
\end{equation}
where $z\in\mathcal{Z}$ denotes the spatial index, and $\vert\mathcal{Z}\vert$ is the number of all spatial indices. $\bm y $ indicates the ground-truth HR video sequence and $\epsilon$ is a smoothing parameter set to 1$\times$10$^{-9}$ in our experiments.

\begin{table}[!t]
\centering
\setlength{\abovecaptionskip}{0pt} 
\setlength{\belowcaptionskip}{0pt}
\caption{Quantitative comparison with state-of-the-art methods for {AVSR} on the Vid4 dataset (PSNR$\uparrow$ / SSIM$\uparrow$ / LPIPS$\downarrow$). 
The inference time is averaged over all frames from the four test videos for $\times 4$ SR. }
\label{table:vid4}
\renewcommand{\arraystretch}{1.2}
\resizebox{\columnwidth}{!}{
\begin{tabular}{cc|c|c|c|c}
\multicolumn{2}{c|}{Method} & \multicolumn{3}{|c}{Scale} & \multicolumn{1}{|c}{\multirow{3}{*}{Inference time (s)}} \\ 
\cline{1-5}
\multirow{2}{*}{Backbone} & Upsampling  & \multirow{2}{*}{$\times \frac{2.5}{3.5}$} & \multirow{2}{*}{$\times \frac{4}{4}$} & \multirow{2}{*}{$\times \frac{7.2}{6}$}  &\\ 
&  Unit & & & &  \\ \hline 
\multicolumn{2}{c|}{Bicubic} & {\fontsize{8}{10}\selectfont 23.00/0.728/0.396 } & {\fontsize{8}{10}\selectfont 20.96/0.617/0.498} & {\fontsize{8}{10}\selectfont 18.73/0.463/0.691} &  {\fontsize{8}{10}\selectfont ---} \\ 
\multicolumn{2}{c|}{ArbSR~\cite{wang2021learning}} & {\fontsize{8}{10}\selectfont 25.86/0.815/0.224} & {\fontsize{8}{10}\selectfont 24.01/0.721/0.313 } &  {\fontsize{8}{10}\selectfont 21.23/0.540/0.478 } & {\fontsize{8}{10}\selectfont  0.2955 }   \\
\multicolumn{2}{c|}{EQSR~\cite{wang2023deep}}   & {\fontsize{8}{10}\selectfont  26.24/0.826/0.210 }  & {\fontsize{8}{10}\selectfont 24.16/0.730/0.300 }    & {\fontsize{7.5}{9}\selectfont \textbf{21.72}/0.573/0.443 }        & {\fontsize{8}{10}\selectfont  0.4181  }      \\ \hline 
\multirow{3}{*}{RDN~\cite{zhang2018residual}} & LTE~\cite{lee2022local} & {\fontsize{8}{10}\selectfont  25.98/0.818/0.226 }  & {\fontsize{8}{10}\selectfont  24.03/0.722/0.312  }     & {\fontsize{8}{10}\selectfont  21.64/0.565/0.455 }     & {\fontsize{8}{10}\selectfont 0.2363  }    \\ 
& CLIT~\cite{chen2023cascaded}  & {\fontsize{8}{10}\selectfont  25.83/0.815/0.223 }   & {\fontsize{8}{10}\selectfont  23.94/0.721/0.312 }      & {\fontsize{8}{10}\selectfont  21.62/0.563/0.458 }           & {\fontsize{8}{10}\selectfont  0.7805   }   \\ 
& OPE~\cite{song2023ope}  & {\fontsize{8}{10}\selectfont 25.77/0.818/0.217 }  & {\fontsize{8}{10}\selectfont 23.98/0.719/0.317 }    & {\fontsize{8}{10}\selectfont 21.60/0.559/0.483 }   & {\fontsize{8}{10}\selectfont 0.1242 }   \\ \hline 
\multirow{3}{*}{SwinIR~\cite{liang2021swinir}} & LTE~\cite{lee2022local}    & {\fontsize{8}{10}\selectfont 26.43/0.826/0.217}   & {\fontsize{8}{10}\selectfont  24.09/0.727/0.305  }     & {\fontsize{7.5}{9}\selectfont \textbf{21.72}/0.570/0.448 }      & {\fontsize{8}{10}\selectfont 0.3332  }         \\ 
& CLIT~\cite{chen2023cascaded}  & {\fontsize{8}{10}\selectfont  25.89/0.818/0.224 }   & {\fontsize{8}{10}\selectfont  24.00/0.724/0.314 }         & {\fontsize{8}{10}\selectfont 21.65/0.565/0.457 }     & {\fontsize{8}{10}\selectfont  0.9016  }  \\ 
& OPE~\cite{song2023ope}   & {\fontsize{8}{10}\selectfont 25.55/0.801/0.221 } & {\fontsize{8}{10}\selectfont  23.93/0.711/0.320 }     & {\fontsize{8}{10}\selectfont 21.58/0.551/0.471 }          & {\fontsize{8}{10}\selectfont 0.2008 }    \\ \hline 
\multicolumn{2}{c|}{VideoINR~\cite{chen2022videoinr}}    &  {\fontsize{8}{10}\selectfont 23.02/0.715/0.203}  &  {\fontsize{8}{10}\selectfont 24.34/0.741/0.249 }    & {\fontsize{8}{10}\selectfont 20.80/0.536/0.431 }   & {\fontsize{8}{10}\selectfont  0.2364 }  \\ 
\multicolumn{2}{c|}{MoTIF~\cite{chen2023motif}}    & {\fontsize{8}{10}\selectfont 23.55/0.734/0.209  }    & {\fontsize{8}{10}\selectfont 24.52/0.746/0.261 }      & {\fontsize{8}{10}\selectfont 20.94/0.546/0.426 }    & {\fontsize{8}{10}\selectfont 0.4053 }     \\  \hline
\multicolumn{2}{c|}{ST-AVSR (Ours)}    & {\fontsize{7}{9}\selectfont  \textbf{29.09}/\textbf{0.913}/\textbf{0.069}}     & {\fontsize{7}{9}\selectfont \textbf{26.16}/\textbf{0.852}/\textbf{0.127}}       & {\fontsize{7.5}{9}\selectfont 21.60/\textbf{0.668}/\textbf{0.306} }      & {\fontsize{8}{10}\selectfont \textbf{0.0495} }   
\\ 
\end{tabular}}
\end{table}

\begin{figure*}[!t]\footnotesize
	\centering
	\setlength{\abovecaptionskip}{3pt} 
	\setlength{\belowcaptionskip}{0pt}
	\begin{tabular}{l}
		\includegraphics[width=0.95\linewidth]{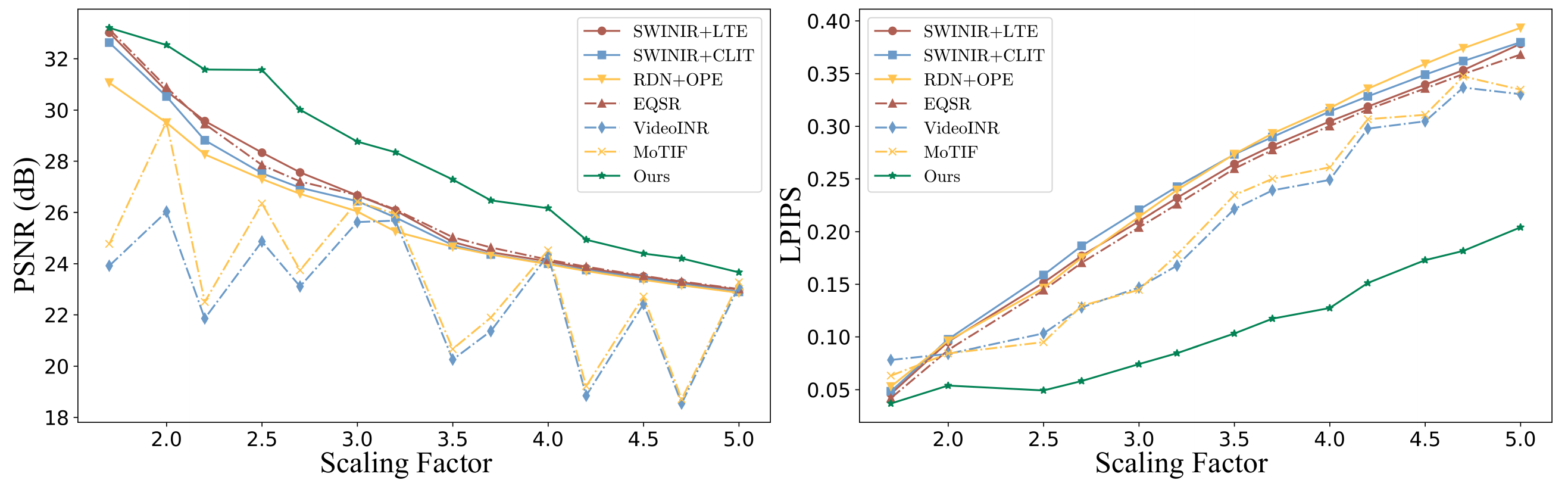}\\
	\end{tabular}
	\caption{
		PSNR and LPIPS variations for different scaling factors on Vid4.
	}
	\label{fig:plot}
\end{figure*}
 
\begin{figure*}[!t]\footnotesize
	\centering
	\setlength{\abovecaptionskip}{3pt} 
	\setlength{\belowcaptionskip}{0pt}
	\begin{tabular}{cccccc}
		\includegraphics[width=\linewidth]{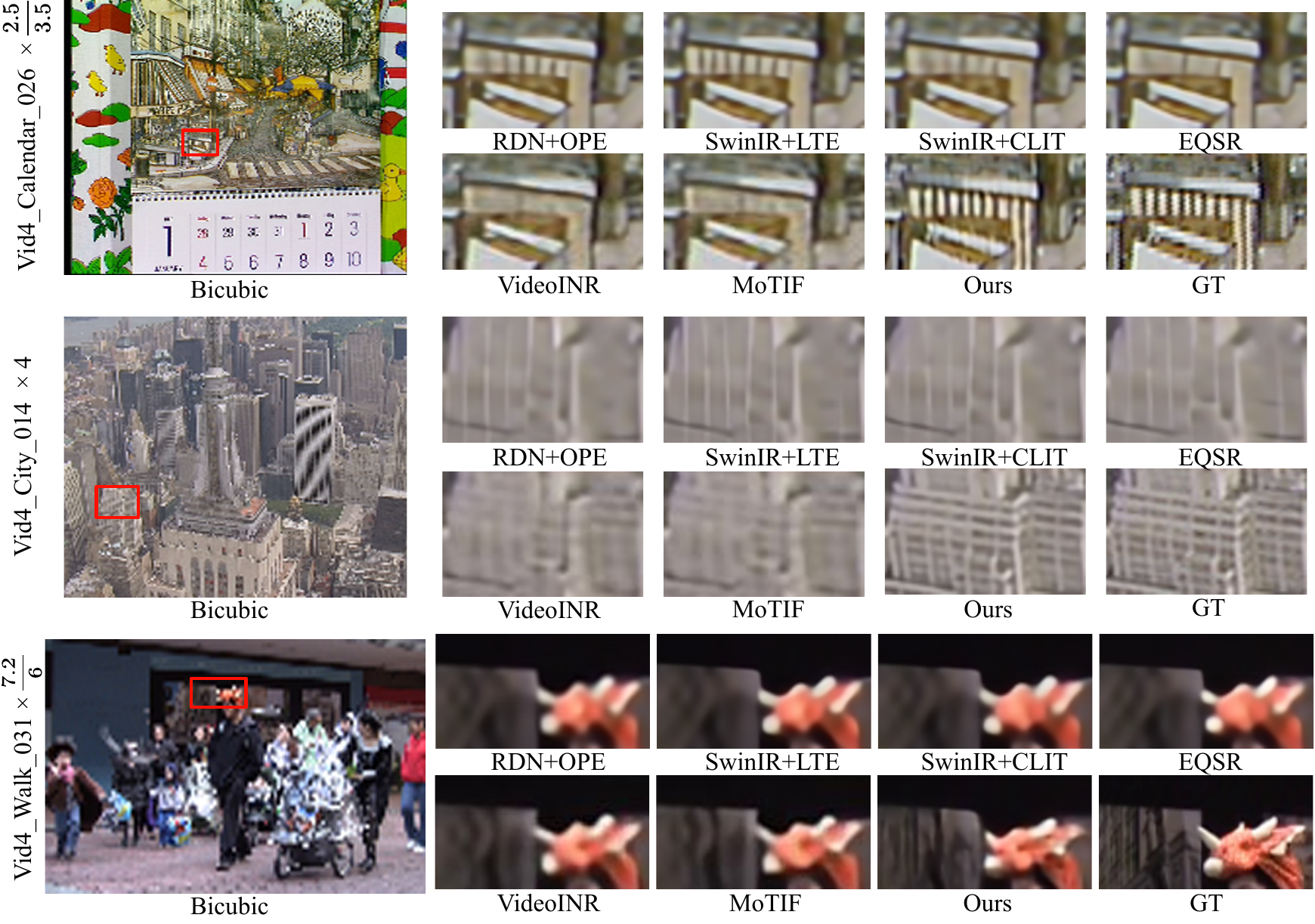}\\
	\end{tabular}
	\caption{
		Visual comparison of different AVSR methods on Vid4.
	}
	\label{fig:vid4}
\end{figure*}
\subsection{Comparison with State-of-the-art Methods}
We compare ST-AVSR with state-of-the-art AISR and AVSR methods.
For AISR, we choose methods from two categories: 1) learnable adaptive filter-based upsampling, including ArbSR~\cite{wang2021learning} and EQSR~\cite{wang2023deep} and 2) implicit neural representation-based upsampling, including LTE~\cite{lee2022local}, CLIT~\cite{chen2023cascaded} and OPE~\cite{song2023ope}.
For AVSR, we compare with VideoINR~\cite{chen2022videoinr} and MoTIF~\cite{chen2023motif}.
Additionally, we include EDVR~\cite{wang2019edvr}, a state-of-the-art VSR method for integer scaling factors.   
All competing methods have been finetuned on the REDS dataset for a fair comparison, and we evaluate their generalization ability on Vid4~\cite{liu2013bayesian} and using unseen degradation models.

\subsubsection{Comparison on REDS.}
Benefiting from the long-term spatiotemporal dependency modeling and the multi-scale structural and textural prior,
ST-AVSR achieves the best results under all evaluation metrics and across all scaling factors, presented in Table~\ref{table:reds}. The dramatic visual quality improvements can also be clearly seen in Fig.~\ref{fig:reds}, in which ST-AVSR recovers more faithful detail with less severe distortion across different scales. 
\begin{figure*}[!t]\footnotesize
	\centering
	\setlength{\abovecaptionskip}{3pt} 
	\setlength{\belowcaptionskip}{0pt}
	\begin{tabular}{cccccc}
		\includegraphics[width=\linewidth]{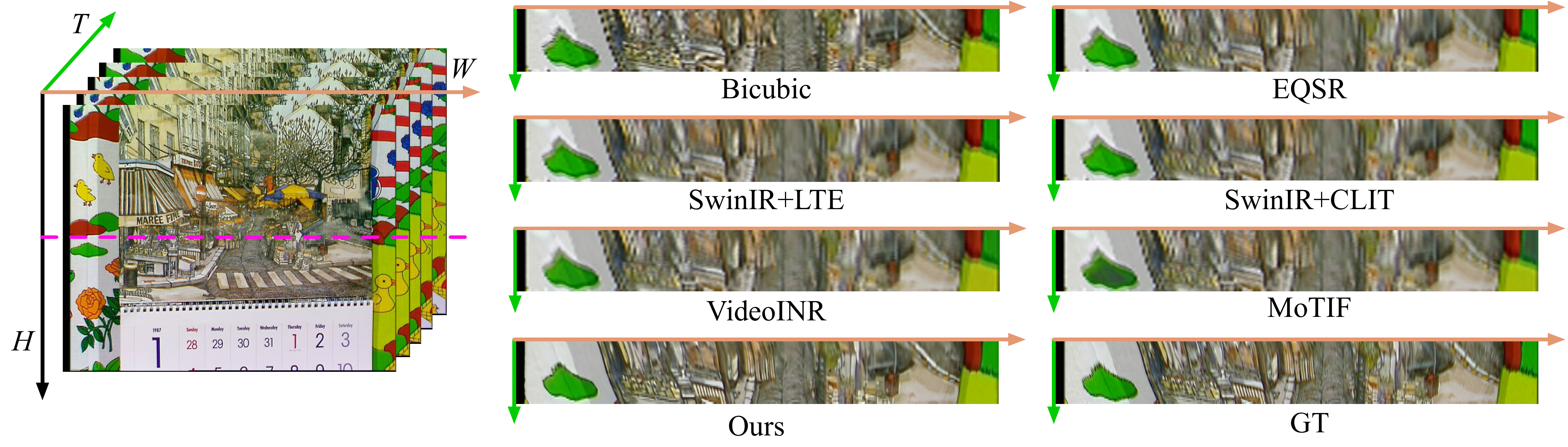}\\
	\end{tabular}
	\caption{
		Temporal consistency comparison.
		We visualize the pixel variations in the row indicated by the pink dashed line along the temporal dimension. 
	}
	\label{fig:temporal}
\end{figure*}

\subsubsection{Generalization on Vid4.}
AVSR models trained on REDS are directly applicable to Vid4, which serves as a generalization test.
The quantitative results, listed in Table~\ref{table:vid4}, indicate that 
 ST-AVSR surpasses all competing methods by wide margins in terms of SSIM and LPIPS  across varying scaling factors. A closer look is provided in Fig.~\ref{fig:plot}, illustrating the PSNR and LPIPS variations for different scaling factors. 
It is evident that existing AVSR methods, particularly VideoINR and MoTIF, fail to achieve satisfactory SR performance for non-integer and asymmetric scales. This issue is mainly due to the pixel misalignment between the super-resolved and ground-truth frames,  leading to oscillating PSNR values. Such oscillation is less pronounced in terms of LPIPS as it offers some degree of robustness to misalignment through the VGG feature hierarchy. As for ST-AVSR, it degrades gracefully with increasing scaling factors, including non-integer and asymmetric ones. 
Table~\ref{table:vid4} also presents the average inference time for each competing method, measured over all frames from Vid4 for $\times 4$ SR using an NVIDIA RTX A6000 GPU. ST-AVSR runs nearly in real-time and is significantly faster than all competing methods, especially those based on implicit neural representations.
\begin{figure*}[t]\footnotesize
    \centering
    \setlength{\abovecaptionskip}{3pt} 
    \setlength{\belowcaptionskip}{0pt}
    \begin{tabular}{cccc}
        \includegraphics[width=1\linewidth]{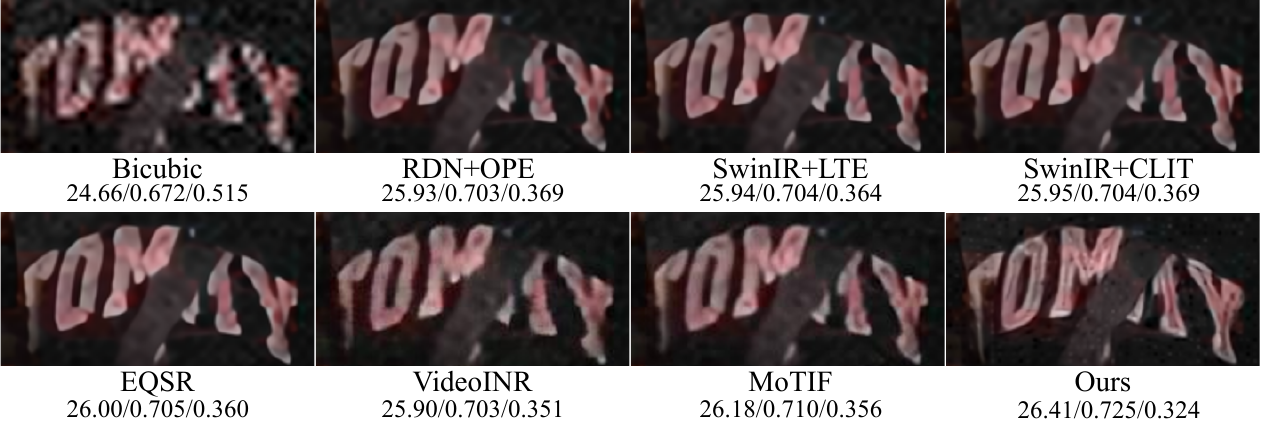}\\
    \end{tabular}
    \caption{Qualitative and quantitative (PSNR$\uparrow$ / SSIM$\uparrow$ / LPIPS$\downarrow$) comparison of different AVSR methods under an unseen degradation model. 
    }
    \label{fig:degrad}
\end{figure*}

Qualitative results are shown in Fig.~\ref{fig:vid4}, where we find that ST-AVSR 
consistently produces natural and visually pleasing SR outputs. It excels in reconstructing both non-structured and structured texture, which we believe arises from the incorporation of the multi-scale structural and textural prior, as also supported by previous studies~\cite{ding2021comparison}. 
Additionally,  Fig.~\ref{fig:temporal} compares temporal consistency by unfolding one row of pixels as indicated by the pink dashed line along the temporal dimension. The temporal profiles of the competing methods appear blurry and zigzagging, indicating temporal flickering artifacts. In contrast, the temporal profile of ST-AVSR is closer to the ground-truth, with a sharper and smoother visual appearance.
\subsubsection{Generalization to Unseen Degradation Models.} A practical AVSR method must be effective under various, potentially unseen degradation models. To evaluate this, we generate test video sequences by incorporating more complex video degradations~\cite{chan2022investigating}, such as noise contamination and video compression before bicubic downsampling, which are absent from the training data. Fig.~\ref{fig:degrad} presents visual comparison of $\times 4$ SR results. Given the degradation gap between training and testing, all methods, including ST-AVSR, exhibit some form of artifacts. Nevertheless, the result by ST-AVSR appears more natural and less distorted, as also confirmed by higher objective quality values.

\begin{table}[!t]
\centering
\setlength{\abovecaptionskip}{0pt} 
\setlength{\belowcaptionskip}{0pt}
\caption{Ablation analysis of ST-AVSR on REDS (PSNR$\uparrow$ / SSIM$\uparrow$ / LPIPS$\downarrow$). See the text for the details of different variants.}
\label{table:ablation}
\resizebox{\columnwidth}{!}{
\begin{tabular}{cc|c|c|c|c|c}
 \multirow{2}{*}{} & \multirow{2}{*}{} & \multicolumn{5}{|c}{Scale} \\  
\cline{3-7}
 & & {$\times 2$} & {$\times 3$} & {$\times 4$} & {$\times 6$} & {$\times 8$}  \\ 
\hline 
\multicolumn{2}{c|}{Variant 1)} & {\fontsize{8}{10}\selectfont 35.23/0.952/0.048 }  & {\fontsize{8}{10}\selectfont 31.56/0.907/0.117}     & {\fontsize{8}{10}\selectfont 29.31/0.851/0.168}     & {\fontsize{8}{10}\selectfont 26.63/0.770/0.271}    & {\fontsize{8}{10}\selectfont 25.09/0.701/0.339 }   \\
\multicolumn{2}{c|}{Variant 2)} & {\fontsize{8}{10}\selectfont 36.74/0.968/0.043} & {\fontsize{8}{10}\selectfont 33.02/0.932/0.072} & {\fontsize{8}{10}\selectfont 30.68/0.889/0.124}&{\fontsize{8}{10}\selectfont 27.57/0.801/0.233} & {\fontsize{8}{10}\selectfont 25.76/0.735/0.305} \\
\multicolumn{2}{c|}{Variant 3)}  &  {\fontsize{8}{10}\selectfont 36.39/0.965/0.043} &  {\fontsize{8}{10}\selectfont 32.65/0.927/0.080} &  {\fontsize{8}{10}\selectfont 30.39/0.883/0.133}  &  {\fontsize{8}{10}\selectfont 27.43/0.796/0.240}  & {\fontsize{8}{10}\selectfont 25.64/0.730/0.313}   \\
\multicolumn{2}{c|}{Variant 4)}  & {\fontsize{8}{10}\selectfont 36.62/0.967/0.040} & {\fontsize{8}{10}\selectfont 32.75/0.928/0.077} & {\fontsize{8}{10}\selectfont 30.36/0.882/0.131}&{\fontsize{8}{10}\selectfont 27.29/0.790/0.239}& {\fontsize{8}{10}\selectfont 25.47/0.722/0.314} \\
\multicolumn{2}{c|}{Variant 5)} & {\fontsize{8}{10}\selectfont 36.48/0.966/0.039} & {\fontsize{8}{10}\selectfont 32.78/0.928/0.074} & {\fontsize{8}{10}\selectfont 30.41/0.883/0.127}&{\fontsize{8}{10}\selectfont 27.37/0.793/0.239}& {\fontsize{8}{10}\selectfont 25.59/0.727/0.316} \\
\multicolumn{2}{c|}{Variant 6)} & {\fontsize{8}{10}\selectfont 36.34/0.965/0.044} & {\fontsize{8}{10}\selectfont 32.64/0.927/0.077} & {\fontsize{8}{10}\selectfont 30.29/0.881/0.131}&{\fontsize{8}{10}\selectfont 27.28/0.792/0.242}& {\fontsize{8}{10}\selectfont 25.50/0.725/0.318} \\
\hline 
\multicolumn{2}{c|}{B-AVSR}  & {\fontsize{8}{10}\selectfont 35.94/0.960/0.058} & {\fontsize{8}{10}\selectfont 31.86/0.910/0.110} & {\fontsize{8}{10}\selectfont 29.67/0.861/0.168}&{\fontsize{8}{10}\selectfont 26.83/0.771/0.269}& {\fontsize{8}{10}\selectfont 25.13/0.706/0.339} \\
\multicolumn{2}{c|}{ST-AVSR ($L=0$)} & {\fontsize{8}{10}\selectfont 36.15/0.963/0.047} &  {\fontsize{8}{10}\selectfont 32.42/0.924/0.080}& {\fontsize{8}{10}\selectfont 30.12/0.879/0.135}&{\fontsize{8}{10}\selectfont 27.18/0.790/0.249} & {\fontsize{8}{10}\selectfont 25.44/0.725/0.323} \\
\multicolumn{2}{c|}{ST-AVSR ($L=1$)} & {\fontsize{8}{10}\selectfont 36.44/0.966/0.045} &  {\fontsize{8}{10}\selectfont 32.93/0.929/0.079}& {\fontsize{8}{10}\selectfont 30.49/0.883/0.131}& {\fontsize{8}{10}\selectfont 27.39/0.796/0.231}& {\fontsize{8}{10}\selectfont 25.60/0.729/0.313} \\
\multicolumn{2}{c|}{ST-AVSR ($L=2$)}   & {\fontsize{8}{10}\selectfont 36.91/0.969/0.041}    & {\fontsize{8}{10}\selectfont  33.41/0.937/0.066}       & {\fontsize{7.5}{9}\selectfont  31.03/0.897/\textbf{0.114}}      & {\fontsize{7}{9}\selectfont \textbf{27.89}/\textbf{0.812}/\textbf{0.222} }    & {\fontsize{7}{9}\selectfont \textbf{26.04}/\textbf{0.746}/\textbf{0.298}}     \\
\multicolumn{2}{c|}{ST-AVSR ($L=3$)} & {\fontsize{7}{9}\selectfont \textbf{36.94}/\textbf{0.971}/\textbf{0.040}} & {\fontsize{7}{9}\selectfont \textbf{33.48}/\textbf{0.939}/\textbf{0.065}} & {\fontsize{7}{9}\selectfont \textbf{31.05}/\textbf{0.898}/\textbf{0.114}}& {\fontsize{8}{10}\selectfont 27.88/0.809/0.225}&  {\fontsize{8}{10}\selectfont 26.00/0.740/0.308}\\
\end{tabular}}
\end{table}
\subsection{Ablation Studies} 
We conduct a series of ablation experiments on the flow-refined cross-attention unit, investigating the following variants: 1) disabling the entire unit, 2) disabling flow rectification (Eqs.~\eqref{eq:fr1} and~\eqref{eq:fr2}), 3) disabling flow estimation (\ie, using only deformable convolution as a form of coarse flow estimation), 4) replacing local cross-attention and global self-attention (Eqs.~\eqref{eq:fr3} to~\eqref{eq:fr6}) with na\"{i}ve feature concatenation, 5) disabling only local cross-attention (Eqs.~\eqref{eq:fr3} to~\eqref{eq:fr4}), 6) disabling only global self-attention (Eqs.~\eqref{eq:fr5} and~\eqref{eq:fr6}). We also compare B-AVSR (without the multi-scale structural and textural prior) to ST-AVSR, and vary the length of the local window $L$ in ST-AVSR. The results are shown in Table~\ref{table:ablation}, where we find that all design choices contribute positively to AVSR. Notably, adding the multi-scale structural and textural prior significantly boosts performance by up to $1.5$ dB. Additionally, ST-AVSR benefits from a larger window size to attend to more future frames.
 However, this also increases the computational complexity, and therefore, we set $L=2$ as a reasonable compromise.

\section{Conclusion}
We have introduced an arbitrary-scale video super-resolution method. Our baseline model, B-AVSR, adopts a flow-guided recurrent unit and a flow-refined cross-attention unit to extract, align, and aggregate spatiotemporal features, along with a hyper-upsampling unit for efficient arbitrary-scale upsampling. Furthermore, our complete model, ST-AVSR, integrates a multi-scale structural and textural prior derived from the pre-trained VGG network. Experimental results demonstrate that ST-AVSR outperforms state-of-the-art methods in terms of SR quality, generalization ability, and inference speed.  
In future work, we plan to augment our spatial structural and textural prior with temporal information, and extend ST-AVSR for space-time AVSR.

\section*{Acknowledgements}
This work was supported in part by the National Key Research and Development Program of China (2023YFE0210700), the Hong Kong ITC Innovation and Technology Fund (9440390), the National Natural Science Foundation of China(62172127, 62071407, U22B2035, 62311530101, 62132006), and the Natural Science Foundation of Heilongjiang Province (YQ2022F004).

\bibliographystyle{splncs04}
\bibliography{egbib}

\begin{thebibliography}{10}
\providecommand{\url}[1]{\texttt{#1}}
\providecommand{\urlprefix}{URL }
\providecommand{\doi}[1]{https://doi.org/#1}

\bibitem{behjati2021overnet}
Behjati, P., Rodriguez, P., Mehri, A., Hupont, I., Tena, C.F., Gonzalez, J.:
  Over{N}et: {L}ightweight multi-scale super-resolution with overscaling
  network. In: WACV. pp. 2694--2703 (2021)

\bibitem{caballero2017real}
Caballero, J., Ledig, C., Aitken, A., Acosta, A., Totz, J., Wang, Z., Shi, W.:
  Real-time video super-resolution with spatio-temporal networks and motion
  compensation. In: CVPR. pp. 4778--4787 (2017)

\bibitem{cao2023ciaosr}
Cao, J., Wang, Q., Xian, Y., Li, Y., Ni, B., Pi, Z., Zhang, K., Zhang, Y.,
  Timofte, R., Van~Gool, L.: Ciao{SR}: {C}ontinuous implicit
  attention-in-attention network for arbitrary-scale image super-resolution.
  In: CVPR. pp. 1796--1807 (2023)

\bibitem{chambolle2004algorithm}
Chambolle, A.: An algorithm for total variation minimization and applications.
  JMIV  \textbf{20},  89--97 (2004)

\bibitem{chan2021basicvsr}
Chan, K.C., Wang, X., Yu, K., Dong, C., Loy, C.C.: {BasicVSR}: The search for
  essential components in video super-resolution and beyond. In: CVPR. pp.
  4947--4956 (2021)

\bibitem{chan2022basicvsr++}
Chan, K.C., Zhou, S., Xu, X., Loy, C.C.: {BasicVSR++}: Improving video
  super-resolution with enhanced propagation and alignment. In: CVPR. pp.
  5972--5981 (2022)

\bibitem{chan2022investigating}
Chan, K.C., Zhou, S., Xu, X., Loy, C.C.: Investigating tradeoffs in real-world
  video super-resolution. In: CVPR. pp. 5962--5971 (2022)

\bibitem{chen2023cascaded}
Chen, H.W., Xu, Y.S., Hong, M.F., Tsai, Y.M., Kuo, H.K., Lee, C.Y.: Cascaded
  local implicit transformer for arbitrary-scale super-resolution. In: CVPR.
  pp. 18257--18267 (2023)

\bibitem{chen2023motif}
Chen, Y.H., Chen, S.C., Lin, Y.Y., Peng, W.H.: Mo{TIF}: {L}earning motion
  trajectories with local implicit neural functions for continuous space-time
  video super-resolution. In: ICCV. pp. 23131--23141 (2023)

\bibitem{chen2021learning}
Chen, Y., Liu, S., Wang, X.: Learning continuous image representation with
  local implicit image function. In: CVPR. pp. 8628--8638 (2021)

\bibitem{chen2022videoinr}
Chen, Z., Chen, Y., Liu, J., Xu, X., Goel, V., Wang, Z., Shi, H., Wang, X.:
  Video{INR}: {L}earning video implicit neural representation for continuous
  space-time super-resolution. In: CVPR. pp. 2047--2057 (2022)

\bibitem{ding2021adists}
Ding, K., Liu, Y., Zou, X., Wang, S., Ma, K.: Locally adaptive structure and
  texture similarity for image quality assessment. In: ACMMM. pp. 2483--2491
  (2021)

\bibitem{ding2021comparison}
Ding, K., Ma, K., Wang, S., Simoncelli, E.P.: Comparison of full-reference
  image quality models for optimization of image processing systems. IJCV
  \textbf{129}(4),  1258--1281 (2021)

\bibitem{dong2014learning}
Dong, C., Loy, C.C., He, K., Tang, X.: Learning a deep convolutional network
  for image super-resolution. In: ECCV. pp. 184--199 (2014)

\bibitem{donoho2006compressed}
Donoho, D.L.: Compressed sensing. IEEE TIT  \textbf{52}(4),  1289--1306 (2006)

\bibitem{fu2023dreamsim}
Fu, S., Tamir, N.Y., Sundaram, S., Chai, L., Zhang, R., Dekel, T., Isola, P.:
  Dream{S}im: {L}earning new dimensions of human visual similarity using
  synthetic data. NeurIPS pp. 50742--50768 (2023)

\bibitem{fu2021residual}
Fu, Y., Chen, J., Zhang, T., Lin, Y.: Residual scale attention network for
  arbitrary scale image super-resolution. Neurocomputing  \textbf{427},
  201--211 (2021)

\bibitem{glasner2009super}
Glasner, D., Bagon, S., Irani, M.: Super-resolution from a single image. In:
  ICCV. pp. 349--356 (2009)

\bibitem{ha2017hypernetworks}
Ha, D., Dai, A.M., Le, Q.V.: Hypernetworks. In: ICLR (2017)

\bibitem{he2011single}
He, H., Siu, W.C.: Single image super-resolution using {G}aussian process
  regression. In: CVPR. pp. 449--456 (2011)

\bibitem{hu2018squeeze}
Hu, J., Shen, L., Sun, G.: Squeeze-and-excitation networks. In: CVPR. pp.
  7132--7141 (2018)

\bibitem{hu2019meta}
Hu, X., Mu, H., Zhang, X., Wang, Z., Tan, T., Sun, J.: Meta-{SR}: {A}
  magnification-arbitrary network for super-resolution. In: CVPR. pp.
  1575--1584 (2019)

\bibitem{huxley1997gaussian}
Huxley, T.H., Sporring, J.: Gaussian {S}cale-{S}pace {T}heory. Kluwer
  {A}cademic {P}ublishers (1997)

\bibitem{irani1991improving}
Irani, M., Peleg, S.: Improving resolution by image registration. Graphical
  {M}odels and {I}mage {P}rocessing  \textbf{53}(3),  231--239 (1991)

\bibitem{kappeler2016video}
Kappeler, A., Yoo, S., Dai, Q., Katsaggelos, A.K.: Video super-resolution with
  convolutional neural networks. IEEE TCI  \textbf{2}(2),  109--122 (2016)

\bibitem{kim2016accurate}
Kim, J., Lee, J.K., Lee, K.M.: Accurate image super-resolution using very deep
  convolutional networks. In: CVPR. pp. 1646--1654 (2016)

\bibitem{kingma2014adam}
Kingma, D.P., Ba, J.: Adam: A method for stochastic optimization. In: ICLR
  (2014)

\bibitem{koenderink1984structure}
Koenderink, J.J.: The structure of images. Biological {C}ybernetics
  \textbf{50}(5),  363--370 (1984)

\bibitem{lai2017deep}
Lai, W.S., Huang, J.B., Ahuja, N., Yang, M.H.: Deep {Laplacian} pyramid
  networks for fast and accurate super-resolution. In: CVPR. pp. 624--632
  (2017)

\bibitem{lee2022local}
Lee, J., Jin, K.H.: Local texture estimator for implicit representation
  function. In: CVPR. pp. 1929--1938 (2022)

\bibitem{liang2021swinir}
Liang, J., Cao, J., Sun, G., Zhang, K., Van~Gool, L., Timofte, R.: Swin{IR}:
  Image restoration using swin transformer. In: ICCVW. pp. 1833--1844 (2021)

\bibitem{liang2022recurrent}
Liang, J., Fan, Y., Xiang, X., Ranjan, R., Ilg, E., Green, S., Cao, J., Zhang,
  K., Timofte, R., Van~Gool, L.: Recurrent video restoration transformer with
  guided deformable attention. NeurIPS pp. 378--393 (2022)

\bibitem{lim2017enhanced}
Lim, B., Son, S., Kim, H., Nah, S., Mu~Lee, K.: Enhanced deep residual networks
  for single image super-resolution. In: CVPRW. pp. 136--144 (2017)

\bibitem{lindeberg2013scale}
Lindeberg, T.: Scale-{S}pace {T}heory in {C}omputer {V}ision. Springer
  {S}cience \& {B}usiness {M}edia (2013)

\bibitem{liu2013bayesian}
Liu, C., Sun, D.: On bayesian adaptive video super resolution. IEEE TPAMI
  \textbf{36}(2),  346--360 (2013)

\bibitem{loshchilov2016sgdr}
Loshchilov, I., Hutter, F.: {SGDR}: {S}tochastic gradient descent with warm
  restarts. In: ICLR (2017)

\bibitem{lowe2004distinctive}
Lowe, D.G.: Distinctive image features from scale-invariant keypoints. IJCV
  \textbf{60},  91--110 (2004)

\bibitem{mairal2014sparse}
Mairal, J., Bach, F., Ponce, J., et~al.: Sparse modeling for image and vision
  processing. FTCGV  \textbf{8}(2-3),  85--283 (2014)

\bibitem{michalkiewicz2019implicit}
Michalkiewicz, M., Pontes, J.K., Jack, D., Baktashmotlagh, M., Eriksson, A.:
  Implicit surface representations as layers in neural networks. In: ICCV. pp.
  4743--4752 (2019)

\bibitem{mildenhall2020nerf}
Mildenhall, B., Srinivasan, P., Tancik, M., Barron, J., Ramamoorthi, R., Ng,
  R.: Ne{RF}: {R}epresenting scenes as neural radiance fields for view
  synthesis. In: ECCV (2020)

\bibitem{nah2019ntire}
Nah, S., Baik, S., Hong, S., Moon, G., Son, S., Timofte, R., Mu~Lee, K.:
  {NTIRE} 2019 challenge on video deblurring and super-resolution: {D}ataset
  and study. In: CVPRW. pp.~0--0 (2019)

\bibitem{oppenheim1997signals}
Oppenheim, A.V., Willsky, A.S., Nawab, S.H.: Signals \& {S}ystems. Pearson
  {E}ducaci{\'o}n (1997)

\bibitem{peng2020convolutional}
Peng, S., Niemeyer, M., Mescheder, L., Pollefeys, M., Geiger, A.: Convolutional
  occupancy networks. In: ECCV. pp. 523--540 (2020)

\bibitem{shang2023joint}
Shang, W., Ren, D., Yang, Y., Zhang, H., Ma, K., Zuo, W.: Joint video
  multi-frame interpolation and deblurring under unknown exposure time. In:
  CVPR. pp. 13935--13944 (2023)

\bibitem{shechtman2005space}
Shechtman, E., Caspi, Y., Irani, M.: Space-time super-resolution. IEEE TPAMI
  \textbf{27}(4),  531--545 (2005)

\bibitem{shi2016real}
Shi, W., Caballero, J., Husz{\'a}r, F., Totz, J., Aitken, A.P., Bishop, R.,
  Rueckert, D., Wang, Z.: Real-time single image and video super-resolution
  using an efficient sub-pixel convolutional neural network. In: CVPR. pp.
  1874--1883 (2016)

\bibitem{simonyan2015very}
Simonyan, K., Zisserman, A.: Very deep convolutional networks for large-scale
  image recognition. In: ICLR (2015)

\bibitem{sitzmann2020implicit}
Sitzmann, V., Martel, J., Bergman, A., Lindell, D., Wetzstein, G.: Implicit
  neural representations with periodic activation functions. In: NeurIPS. pp.
  7462--7473 (2020)

\bibitem{song2023ope}
Song, G., Sun, Q., Zhang, L., Su, R., Shi, J., He, Y.: {OPE-SR}: Orthogonal
  position encoding for designing a parameter-free upsampling module in
  arbitrary-scale image super-resolution. In: CVPR. pp. 10009--10020 (2023)

\bibitem{sun2018pwc}
Sun, D., Yang, X., Liu, M.Y., Kautz, J.: {PWC-Net}: {CNN}s for optical flow
  using pyramid, warping, and cost volume. In: CVPR. pp. 8934--8943 (2018)

\bibitem{tao2017detail}
Tao, X., Gao, H., Liao, R., Wang, J., Jia, J.: Detail-revealing deep video
  super-resolution. In: ICCV. pp. 4472--4480 (2017)

\bibitem{ulyanov2018deep}
Ulyanov, D., Vedaldi, A., Lempitsky, V.: Deep image prior. In: CVPR. pp.
  9446--9454 (2018)

\bibitem{vasconcelos2023cuf}
Vasconcelos, C.N., Oztireli, C., Matthews, M., Hashemi, M., Swersky, K.,
  Tagliasacchi, A.: C{UF}: {C}ontinuous upsampling filters. In: CVPR. pp.
  9999--10008 (2023)

\bibitem{wandell1995foundations}
Wandell, B.A.: Foundations of {V}ision. {S}inauer {A}ssociates (1995)

\bibitem{wang2021learning}
Wang, L., Wang, Y., Lin, Z., Yang, J., An, W., Guo, Y.: Learning a single
  network for scale-arbitrary super-resolution. In: ICCV. pp. 4801--4810 (2021)

\bibitem{wang2023deep}
Wang, X., Chen, X., Ni, B., Wang, H., Tong, Z., Liu, Y.: Deep arbitrary-scale
  image super-resolution via scale-equivariance pursuit. In: CVPR. pp.
  1786--1795 (2023)

\bibitem{wang2019edvr}
Wang, X., Chan, K.C., Yu, K., Dong, C., Loy, C.C.: {EDVR}: Video restoration
  with enhanced deformable convolutional networks. In: CVPRW. pp.~0--0 (2019)

\bibitem{wang2018recovering}
Wang, X., Yu, K., Dong, C., Loy, C.C.: Recovering realistic texture in image
  super-resolution by deep spatial feature transform. In: CVPR. pp. 606--615
  (2018)

\bibitem{zhang2023lmr}
Zhang, L., Li, X., He, D., Li, F., Ding, E., Zhang, Z.: {LMR}: A large-scale
  multi-reference dataset for reference-based super-resolution. In: ICCV. pp.
  13118--13127 (2023)

\bibitem{zhang2018residual}
Zhang, Y., Tian, Y., Kong, Y., Zhong, B., Fu, Y.: Residual dense network for
  image super-resolution. In: CVPR. pp. 2472--2481 (2018)

\bibitem{zhu2019deformable}
Zhu, X., Hu, H., Lin, S., Dai, J.: Deformable {C}onv{N}ets v2: More deformable,
  better results. In: CVPR. pp. 9308--9316 (2019)

\end{thebibliography}

\newpage
\appendix
\section{Data Pre-Processing and Training Pipeline}
\label{sec:training}
\begin{figure*}[!t]\footnotesize
	\renewcommand\thefigure{S1}
    \setlength{\abovecaptionskip}{3pt} 
	\setlength{\belowcaptionskip}{0pt}
	\begin{tabular}{cccccc}
		\includegraphics[width=1\linewidth]{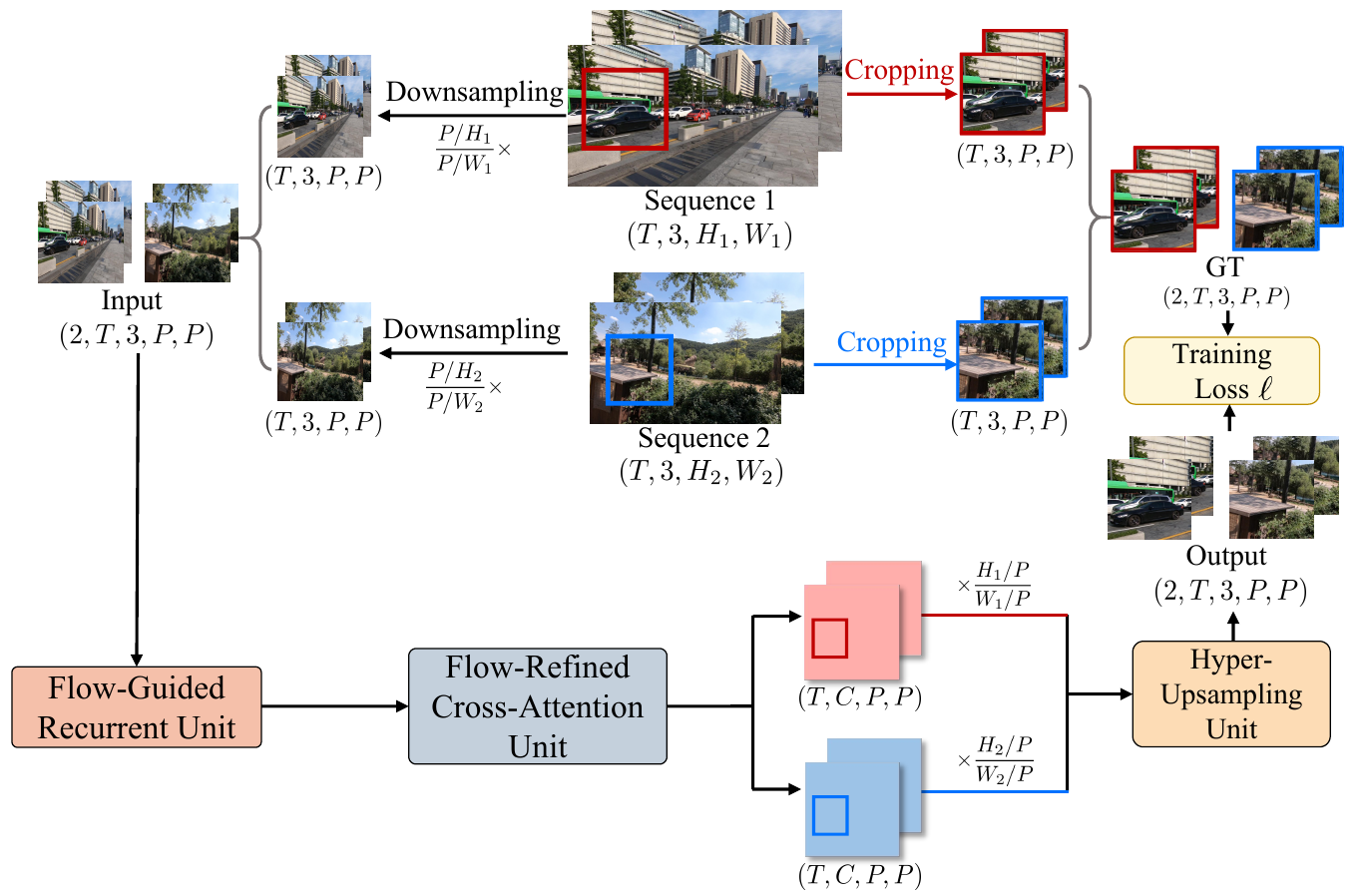}\\
	\end{tabular}
	\caption{Data pre-processing and training pipeline for B-AVSR and ST-AVSR.}
	\label{fig:pipeline}
\end{figure*}

We visualize the data pre-processing and training pipeline in Fig.~\ref{fig:pipeline}, in which we set
 $T=2$.

\section{Visual Comparison of B-AVSR and ST-AVSR}
\label{sec:abla}
\begin{figure*}[!t]\footnotesize
	\renewcommand\thefigure{S2}
	\setlength{\abovecaptionskip}{3pt} 
	\setlength{\belowcaptionskip}{0pt}
	\begin{tabular}{cccccc}
		\includegraphics[width=1\linewidth]{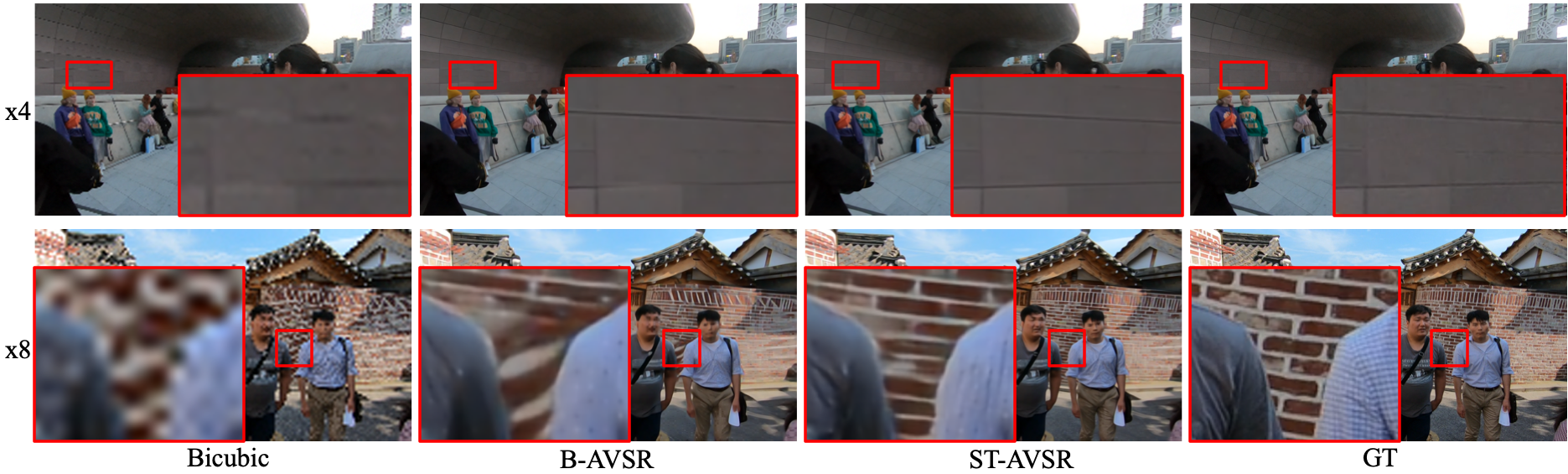}\\
	\end{tabular}
	\caption{Effectiveness of our multi-scale structural and textural prior in aiding AVSR.}
	\label{fig:ab_priors}
\end{figure*}

In the main text, we have provided quantitative comparison of B-AVSR and ST-AVSR. In Fig.~\ref{fig:ab_priors}, we present qualitative comparison, where we observe that our multi-scale structural and textural prior encourages more faithful detail at various scales to be recovered.

\section{More Results on the REDS Dataset}
\begin{figure*}[!t]\footnotesize
	\renewcommand\thefigure{S3}
	\setlength{\abovecaptionskip}{3pt} 
	\setlength{\belowcaptionskip}{0pt}
	\begin{tabular}{cccccc}
		\includegraphics[width=1\linewidth]{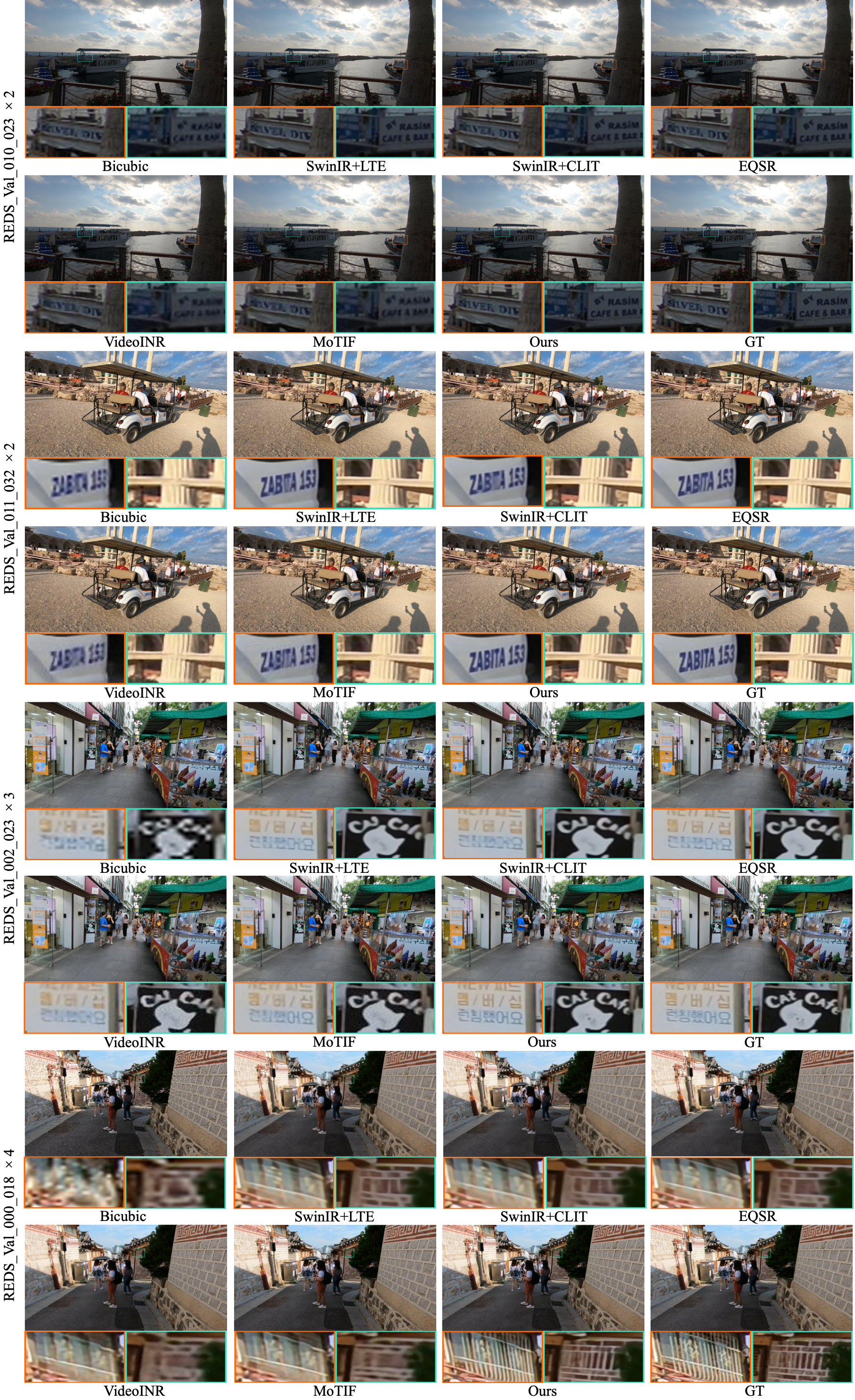}\\
	\end{tabular}
	\caption{More visual results on the REDS dataset~\cite{nah2019ntire}.}
	\label{fig:reds1}
\end{figure*}

\begin{figure*}[!t]\footnotesize
	\renewcommand\thefigure{S4}
	\setlength{\abovecaptionskip}{3pt} 
	\setlength{\belowcaptionskip}{0pt}
	\begin{tabular}{cccccc}
		\includegraphics[width=1\linewidth]{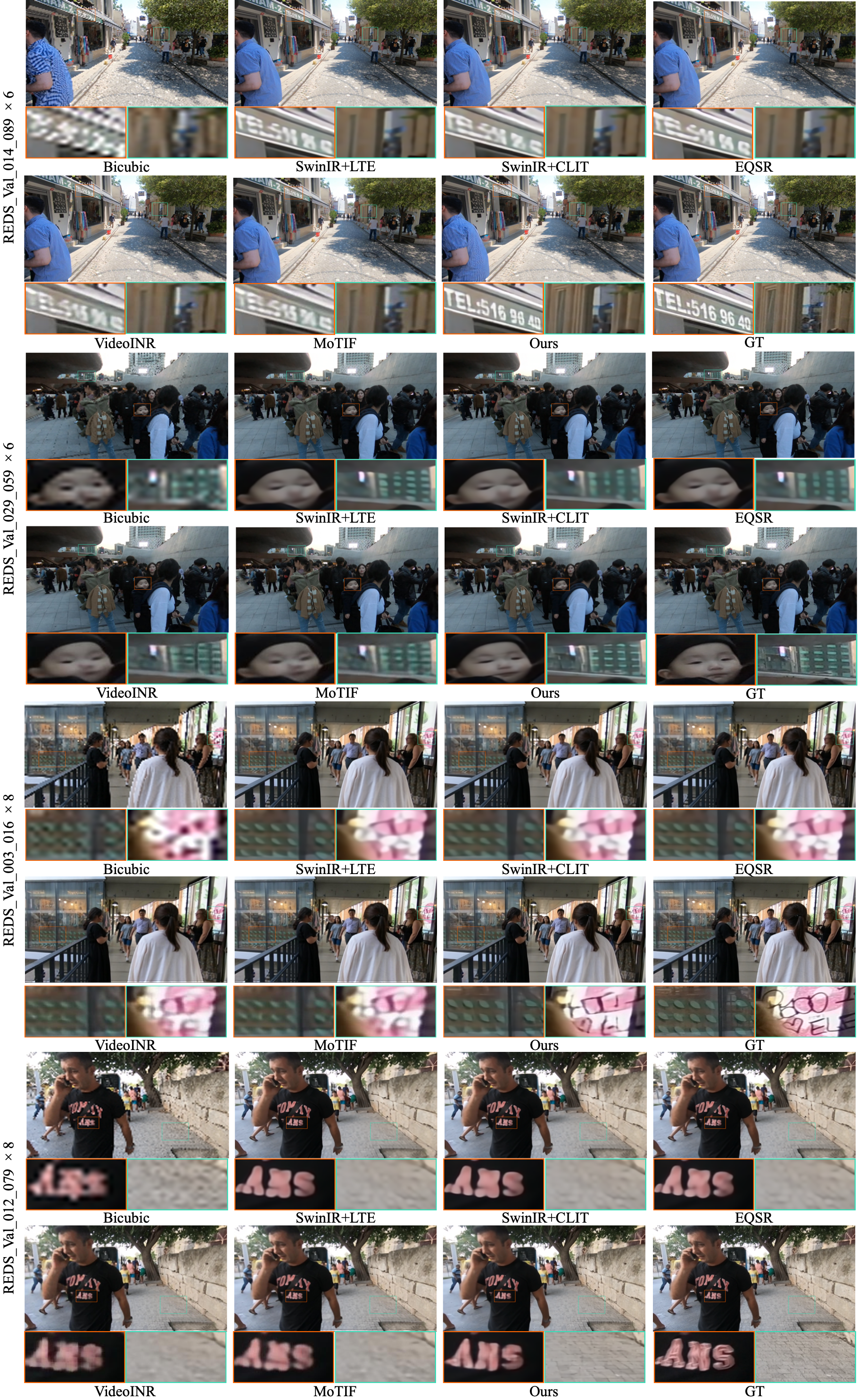}\\
	\end{tabular}
	\caption{More visual results on the REDS dataset~\cite{nah2019ntire}.}
	\label{fig:reds2}
\end{figure*}

We provide more visual results on the REDS dataset in Figs.~\ref{fig:reds1} and~\ref{fig:reds2}.
\label{sec:reds}

\section{More Results on the Vid4 Dataset}
\label{sec:vid4}
\begin{figure*}[!t]\footnotesize
	\renewcommand\thefigure{S5}
    \setlength{\abovecaptionskip}{3pt} 
	\setlength{\belowcaptionskip}{0pt}
	\begin{tabular}{cccccc}
		\includegraphics[width=1\linewidth]{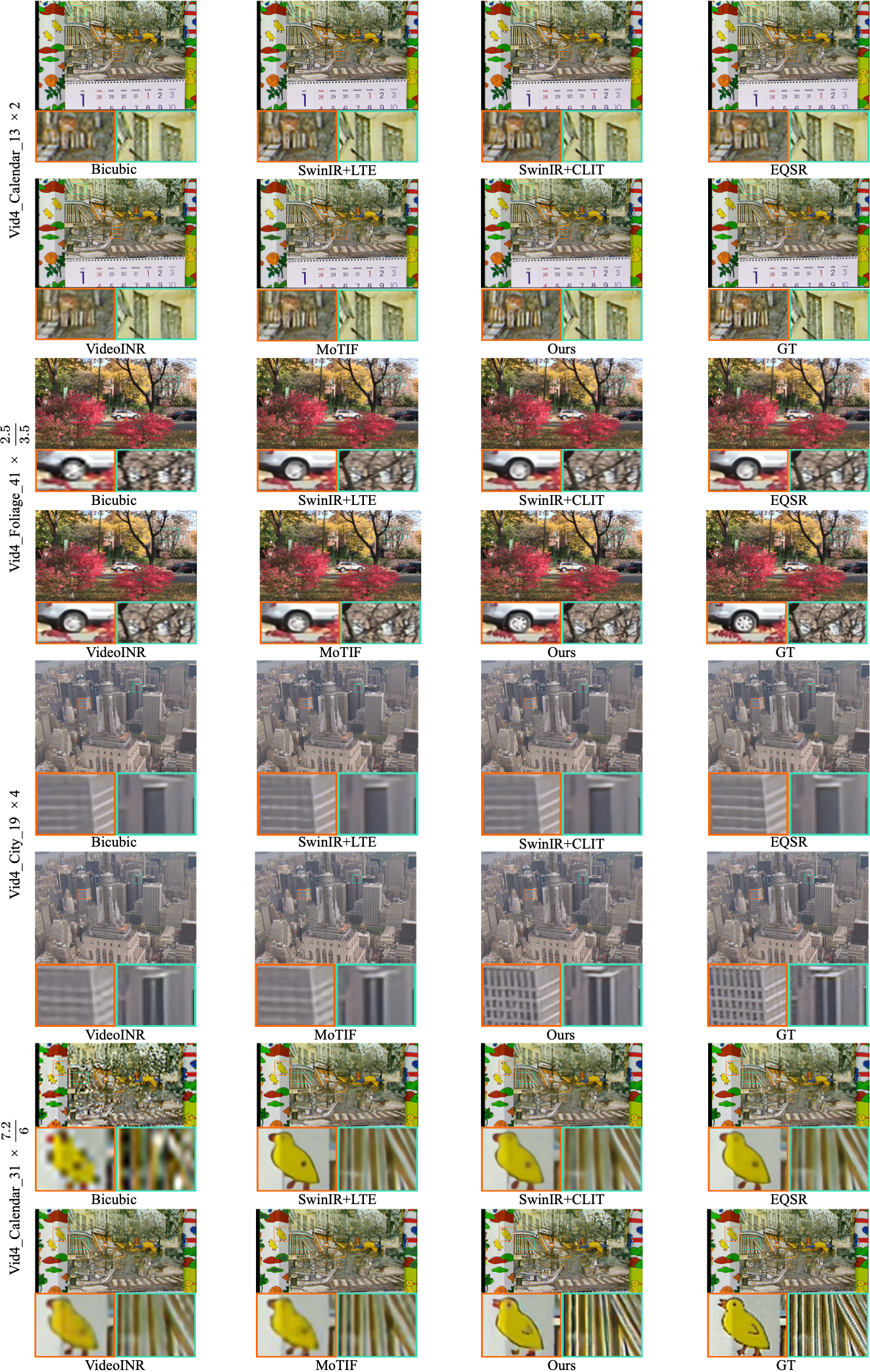}\\
	\end{tabular}
	\caption{More visual results on the Vid4 dataset~\cite{liu2013bayesian}.}
	\label{fig:Vid1}
\end{figure*}

We provide more visual results on the Vid4 dataset in Fig.~\ref{fig:Vid1}.

\end{sloppypar}
\end{document}